\documentclass[12pt,a4paper]{article}
\usepackage{fullpage}
\usepackage[authoryear,round]{natbib}
\usepackage{amsmath}
\usepackage{amsfonts}
\usepackage{amssymb}
\usepackage{graphicx}
\usepackage{amsthm}
\usepackage{rotating}
\usepackage{multirow}

\newtheorem{thm}{Theorem}

\newtheorem{lem}{Lemma}
\newcommand*{\QEDB}{\hfill\ensuremath{\square}}

\title{Theoretical and Experimental Analyses of
  Tensor-Based Regression and Classification}

\author{
Kishan Wimalawarne \\
Department of Computer Science\\
Tokyo Institute of Technology\\
2-12-1 Ookayama, Meguro-ku,\\
Tokyo 152-8552, Japan.\\
\texttt{kishanwn@gmail.com}
\and Ryota Tomioka\\
The Toyota Technological Institute at Chicago,\\
6045 S. Kenwood Av., \\
Chicago, IL 60637, USA. \\
\texttt{tomioka@ttic.edu}  
\and Masashi Sugiyama\\
Department of Complexity Science and Engineering\\
University of Tokyo\\
7-3-1 Hongo, Bunkyo-ku,\\
Tokyo 113-0033, Japan.\\
\texttt{sugi@k.u-tokyo.ac.jp} 
}

\date{}
\begin{document}
\maketitle

\begin{abstract}\noindent
We theoretically and experimentally
investigate tensor-based regression and classification.
Our focus is regularization with 
various tensor norms, including
the \emph{overlapped trace norm},
the \emph{latent trace norm},
and the \emph{scaled latent trace norm}.
We first give dual optimization methods using the \emph{alternating direction method of multipliers},
which is computationally efficient when the number of training samples is moderate.
We then theoretically derive an excess risk bound for each tensor norm
and clarify their behavior.
Finally, we perform extensive experiments using simulated and real data
and demonstrate the superiority of tensor-based learning methods
over vector- and matrix-based learning methods.
\end{abstract}

\section{Introduction}
A wide range of real-world data takes the format of \emph{matrices} and \emph{tensors}, e.g.,
recommendation \citep{Karatzoglou:2010:MRN:1864708.1864727}, video sequences \citep{Kim07r.:tensor}, climates \citep{NIPS2014_5429}, genomes \citep{10.1371/journal.pone.0121396}, and neuro-imaging \citep{doi:10.1080/01621459.2013.776499}.
A naive way to learn from such matrix and tensor data is to vectorize them
and apply ordinary regression or classification methods designed for vectorial data.
However, such a vectorization approach 
would lead to loss in structural information of matrices and tensors
such as \emph{low-rankness}.

The objective of this paper is to 
investigate regression and classification methods
that directly handle tensor data without vectorization.
Low-rank structure of \emph{data} has been
successfully utilized in various applications such as
missing data imputation \citep{cai2010},
robust principal component analysis \citep{Candes:2011:RPC:1970392.1970395},
and subspace clustering \citep{conf/icml/LiuLY10}.
In this paper, instead of low-rankness of data itself,
we consider its dual---\emph{learning coefficients} of a regressor and a classifier.
Low-rankness in learning coefficients means that
only a subspace of feature space is used for regression and classification.

For matrices, regression and classification has been studied in
\citet{conf/icml/TomiokaA07} and
\citet{RSSB:RSSB12031} in the context of EEG data analysis.
It was experimentally demonstrated that
directly learning matrix data
by low-rank regularization can significantly improve the performance
compared to learning after vectorization.
Another advantage of using low-rank regularization
in the context of EEG data analysis is that
analyzing \emph{singular value spectra} of learning coefficients is useful
in understanding activities of brain regions.

More recently, an inductive learning method for tensors has been explored
\citep{SigDinDeLSuy13}.
Compared to the matrix case, learning with tensors is inherently more complex.
For example, the \emph{multilinear} ranks of tensors make it more complicated to
find a proper low-rankness of a tensor compared to matrices which has only one rank.
So far, several tensor norms such as
the \emph{overlapped  trace norm} or the \emph{tensor nuclear norm}
\citep{DBLP:conf/iccv/LiuMWY09}, 
the \emph{latent trace norm} \citep{tomioka/nips13/abs-1303-6370},
and the \emph{scaled latent trace norm} \citep{nips-14} have been proposed
and demonstrated to perform well for various tensor structures.
However, theoretical analysis of tensor learning in inductive learning settings
has not been much investigated yet.
Another challenge in inductive tensor learning is efficient optimization strategies,
since tensor data often has much higher dimensionalities than matrix and vector data.

In this paper, we theoretically and experimentally investigate
tensor-based regression and classification
with regularization by the overlapped trace norm,
the latent trace norm, and the scaled latent trace norm.
We first provide their dual formulations and propose optimization procedures using the \emph{alternating direction method of multipliers} \citep{1886529043},
which is computationally efficient when the number of data samples is moderate.
We then derive an excess risk bound for each tensor regularization,
which allows us to theoretically understand the behavior of
tensor norm regularization.
More specifically, we elucidate
that the excess risk of the overlapped trace norm is bounded with
the average multilinear ranks of each mode,
that of the latent trace norm is bounded with the minimum multilinear rank among all modes,
and that of the scaled latent trace norm is bounded
with the minimum ratio between multilinear ranks and mode dimensions.
Finally, for simulated and real tensor data,
we experimentally investigate the behavior of tensor-based regression and classification methods.
The experimental results are in concordance with our theoretical findings,
and tensor-based learning methods compare favorably with vector- and matrix-based methods.

The remainder of the paper is organized as follows.
In Section~2, we formulate the problem of tensor-based supervised learning
and review 
the overlapped trace norm, the latent trace norm, and the scaled latent trace norm.
In Section~3, we derive dual optimization algorithms
based on the alternating direction method of multipliers.
In Section~4, we theoretically give an excess risk bound for each tensor norm.
In Section~5, we give experimental results on both artificial and real-world data 
and illustrate the advantage of tensor-based learning methods.
Finally, in Section~6, we conclude this paper.

\subsection*{Notation} 
Throughout the paper,
we use standard tensor notation following \citet{journals/siamrev/KoldaB09}.
We represent a $K$-way tensor as $\mathcal{W} \in  \mathbb{R}^{n_{1}\times \cdots \times n_{K}}$ that consists of $N = \prod_{k=1}^{K}n_{k}$ elements. A mode-$k$ fiber of $\mathcal{W}$ is an $n_{k}$-dimensional vector which can be obtained by fixing all except the $k$th index. The mode-$k$ unfolding of  tensor $\mathcal{W}$ is represented as $W_{(k)} \in \mathbb{R}^{n_{k}\times N/n_{k}}$ which is obtained by concatenating all the $N/n_{k}$ mode-$k$ fibers along its columns.   The spectral norm of a matrix $X$ is denoted by $\|X\|_{{\mathrm{op}}} $ which is the maximum singular value of $X$. The operator $ \left \langle \mathcal{W} , \mathcal{X} \right \rangle $ is the sum of element-wise multiplications of $\mathcal{W}$  and $\mathcal{X}$, i.e., $ \left \langle \mathcal{W} , \mathcal{X} \right \rangle = \mathrm{vec}(\mathcal{W})^{\top}\mathrm{vec}(\mathcal{X})$. The Frobenius norm of a tensor $\mathcal{X}$ is defined as $\|\mathcal{X}\|_{{\mathrm{F}}} = \sqrt{ \left \langle \mathcal{X} , \mathcal{X} \right \rangle } $.

\section{Learning with Tensor Regularization}

%In this section we review existing inductive tensor learning models with tensor regularizations and different tensor   norms used for low rank regularization. 

In this section, we put forward inductive tensor learning models with tensor regularization
and review different tensor norms used for low-rank regularization. 

\subsection{Problem Formulation}
Our focus in this paper is regression and classification of tensor data.
Let us consider a data set $(\mathcal{X}_{i}, y_{i}), i = 1,\ldots,m$,
where $\mathcal{X}_{i} \in \mathbb{R}^{n_{1}\times \cdots \times n_{K}}$ is a covariate tensor
and $y_{i}$ is a target.
$y_{i} \in \mathbb{R}$ for regression,
while $y_{i} \in \{-1,1\}$ for classification.
We consider the following learning model for a tensor norm $\| \cdot  \|_{\mathrm{\star}}$:
\begin{equation}
\underset{\mathcal{W},b}{\min} \sum_{i = 1}^{m}l(\mathcal{X}_{i}, y_{i}, \mathcal{W}, b) + \lambda \| \mathcal{W} \|_{\mathrm{\star}} \label{eq:4},
\end{equation}
where $l(\mathcal{X}_{i}, y_{i}, \mathcal{W},b)$ is the loss function:
the squared loss, 
\begin{equation}
l(\mathcal{X}_{i}, y_{i}, \mathcal{W},b) = (y_{i} - ( \left \langle \mathcal{W} ,\mathcal{X}_{i} \right \rangle  + b ) )^{2}, \label{eq:lA}
\end{equation}
is used for regression, and the logistic loss,
\begin{equation}
l(\mathcal{X}_{i}, y_{i}, \mathcal{W},b) = \log(1 + \exp(-y_{i} (\left \langle \mathcal{W}, \mathcal{X}_{i} \right \rangle + b )  ),\label{eq:lB}
\end{equation}
is used for classification.
$b \in \mathbb{R}$ is the bias term and $\lambda \geq 0$ is the regularization parameter. If $\|\cdot\|_{\mathrm{\star}} = \| \cdot \|_{\mathrm{2}}$ or $\| \cdot \|_{\mathrm{1}}$, then the above problem is equivalent to ordinary vector-based $l_{2}$- or $l_{1}$-regularization.

To understand the effect of tensor-based regularization,
it is important to investigate the low-rankness of tensors.
When considering a matrix $W \in \mathbb{R}^{n_{1} \times n_{2}}$,
its \emph{trace norm} is defined as
\begin{equation}
\|W\|_{{\mathrm{tr}}} = \sum_{j=1}^{J} \sigma_{j}, \label{eq:0}
\end{equation}
where $\sigma_{j}$ is the $j^{\mathrm{th}}$ singular value and $J$ is the number of non-zero singular values ($J \leq \min(n_{1},n_{2})$). A matrix is called \emph{law rank} if $J < \min(n_{1},n_{2})$. The matrix trace norm \eqref{eq:0} is a convex envelop to  the matrix rank and it is commonly used
in matrix low-rank approximation \citep{doi:10.1137/070697835}. 

As in matrices, the rank property is also available for tensors, but it is more complicated  due to its multidimensional  structure. The mode-$k$ rank $r_{k}$ of a tensor $\mathcal{W}  \in \mathbb{R}^{n_{1}\times \cdots \times n_{K}}$ is defined as the rank of mode-$k$ unfolding $W_{(k)}$ and the multilinear rank of $\mathcal{W}$ is given as $(r_{1},\ldots,r_{K})$. The mode-$i$ of a tensor $\mathcal{W}$ is called low rank if $r_{i} < n_{i}$.

\subsection{Overlapped Trace Norm}
One of the earliest definitions of a tensor norm is the \emph{tensor nuclear norm} \citep{DBLP:conf/iccv/LiuMWY09} or the \emph{overlapped trace norm} \citep{tomioka/nips13/abs-1303-6370},
which can be represented for a tensor $\mathcal{W} \in \mathbb{R}^{n_{1}\times \cdots \times n_{K}}$ as
\begin{equation}
\| \mathcal{W} \|_{\mathrm{overlap}} = \underset{k=1}{\overset{K}{\sum}} \|W_{(k)}\|_{\mathrm{tr}}.\label{eq:1}
\end{equation}
The overlapped  trace norm can be viewed as a direct extension of the matrix trace norm since it unfolds a tensor on each of its mode and computes the sum of trace norms of the unfolded matrices.  Regularization with the overlapped trace norm can also be seen as an overlapped  group regularization due to the fact that the same tensor is unfolded over different modes and regularized with the trace norm.

 One of the popular applications of the overlapped trace norm is 
\emph{tensor completion} \citep{0266-5611-27-2-025010,DBLP:conf/iccv/LiuMWY09}, where missing entries of a tensor are imputed.  Another application is
\emph{multilinear multitask learning} \citep{conf/icml13/multilmtl},
where multiple vector-based linear learning tasks with a common feature space are arranged as a tensor feature structure and the multiple tasks are solved together with constraints to minimize the multilinear ranks of the tensor feature.

Theoretical analyses on the overlapped norm have been carried out for both tensor completion \citep{tomioka/nips13/abs-1303-6370} and multilinear multitask learning \citep{nips-14};
they have shown that the prediction error of overlapped trace norm regularization is bounded
by the average mode-$k$ ranks which can be large if some modes are close to full rank
even if there are low-rank modes.
Thus, these studies imply that
the overlapped trace norm performs well
when the multilinear ranks have small variations,
and it may result in a poor performance when the multilinear ranks have high variations.

To overcome the weakness of the overlapped trace norm,
recent research in tensor norms has led to new norms such
as the \emph{latent trace norm} \citep{tomioka/nips13/abs-1303-6370}
and the \emph{scaled latent trace norm} \citep{nips-14}.

\subsection{Latent Trace Norm}

\citet{tomioka/nips13/abs-1303-6370} proposed the latent trace norm as
\begin{equation*}
\| \mathcal{W} \|_{\mathrm{latent}} = \underset{\mathcal{W}^{(1)} + \mathcal{W}^{(2)} + \ldots +\mathcal{W}^{(K)} = \mathcal{W} }{\inf} \underset{k=1}{\overset{K}{\sum}}  \| W_{(k)}^{(k)} \|_{\mathrm{tr}}.\label{eq2}
\end{equation*}
The latent trace norm takes a mixture of $K$ latent tensors which is equal to the number of modes,
and regularizes each of them separately. In contrast to the overlapped trace norm, the latent tensor trace norm regularizes different latent tensors for each unfolded mode and this gives the tendency that the latent tensor trace norm  picks the latent tensor with the lowest rank.

 In general, the latent trace norm results in a mixture of latent tensors and the content of each latent tensor would depend on the rank  of its  unfolding.
 In an extreme case, for a tensor with all its modes full except one mode, regularization with the  latent tensor trace norm would result in making the latent tensor  with the lowest mode become prominent while others become  zero.  

\subsection{Scaled Latent Trace Norm}

Recently, \citet{nips-14} proposed the scaled latent trace norm as an extension of the latent trace norm:
\begin{equation*}
\| \mathcal{W} \|_{\mathrm{scaled}} = \underset{\mathcal{W}^{(1)} + \mathcal{W}^{(2)} + \ldots +\mathcal{W}^{(K)} = \mathcal{W} }{\inf} \underset{k=1}{\overset{K}{\sum}} \frac{1}{\sqrt{n_{k}}} \| W_{(k)}^{(k)} \|_{\mathrm{tr}}.\label{eq:3}
\end{equation*}
Compared to the latent trace  norm, the scaled latent trace norm takes the rank relative to the mode dimension.  A major drawback of the latent trace norm is its inability to identify the rank of a mode relative to its dimension. If a tensor has a mode where its dimension is smaller  than other modes yet its relative rank with respect to its mode dimension is high compared  to other modes,
the latent trace norm could incorrectly pick the smallest mode.

The scaled latent norm has the ability to overcome this problem by its scaling with the mode dimensions such that it is able to work with the relative ranks of the tensor. In the context of multilinear multitask learning,
it has been shown that the scaled latent trace norm  works well for tensors
with high variations in multilinear ranks and mode dimensions compared to the overlapped trace norm and the latent trace norm \citep{nips-14}.

The inductive learning setting mentioned in \eqref{eq:4} with the overlapped trace norm has been studied previously in \citet{SigDinDeLSuy13}.
However, theoretical analysis and performance comparison with other tensor norms have not been conducted yet.
Similarly to tensor decomposition \citep{tomioka/nips13/abs-1303-6370} and multilinear multitask learning \citep{nips-14},
tensor-based regression and classification may also be improved
by regularization methods that can work with high variations in multilinear ranks and mode dimensions.

In the following sections,
to make tensor-based learning more practical and to improve the performance,
we consider formulation  \eqref{eq:4} with the overlapped trace norm,
the latent trace norm, and the scaled latent trace norm,
and give computationally efficient optimization algorithms
and excess risk bounds.

\section{Optimization}
%In this section we propose the dual formulation for \eqref{eq:5} and discuss optimization in detail.
In this section,
we consider the dual formulation for \eqref{eq:4} and propose computationally efficient optimization algorithms.
Since optimization of \eqref{eq:4} with regularization using the overlapped trace norm has
already been studied in \citet{SigDinDeLSuy13}, we do not discuss it again here.
Our main focus in this section is optimization of \eqref{eq:4} with regularization 
using the latent trace norm and the scaled latent trace norm. 

Let us consider the formulation \eqref{eq:4} for  a data set
$(\mathcal{X}_{i}, y_{i}) \in \mathbb{R}^{n_{1}\times \cdots \times n_{K}} \times \mathbb{R} , i = 1,\ldots,m$ with latent and scaled latent trace norm regularization as follows:
\begin{equation}
P(\mathcal{W}, b) = \underset{\mathcal{W}^{(1)} + \ldots + \mathcal{W}^{(K)}  = \mathcal{W} , b}{\min} \sum_{i = 1}^{m}l(\mathcal{X}_{i}, y_{i}, \mathcal{W} ,b) + \sum_{k =1}^{K} \lambda_{k} \| W_{(k)}^{(k)} \|_{\mathrm{tr}} \label{eq:5},
\end{equation}
where, for  $k = 1,\ldots,K$ and for any given regularization parameter $\lambda$,
$\lambda_{k} = \lambda$
in the case of the latent trace norm
and $\lambda_{k} = \frac{\lambda}{\sqrt{n_{k}}}$
in the case of the scaled latent trace norm, respectively.
$W_{(k)}^{(k)}$ is the unfolding of  $\mathcal{W}^{(k)}$ on its $k$th mode. It is worth noticing that the application of the latent and scaled latent trace norms requires optimizing over $K$ latent tensors which contain $KN$ variables in total.
For large $K$ and $N$, solving the primal problem \eqref{eq:5} can be computationally expensive
especially in non-linear problems such as logistic regression,
since they require computationally  expensive optimization methods
such as gradient descent or the Newton method.
If the number of training samples $m$ is  $m \ll KN$,
solving the dual problem of \eqref{eq:5} could be computationally more efficient.   
For this reason, we focus on optimization in the dual below.

The dual formulation of \eqref{eq:5} can be written as follows
(its detailed derivation is given in Appendix~A):
\begin{equation*}
\min_{\boldsymbol{\alpha}, \mathcal{V}^{(1)},\cdots, \mathcal{V}^{(K)} } D(-\boldsymbol{\alpha}) +  \sum_{k = 1}^{K} \delta_{\lambda_{k}}(V^{(k)}_{(k)})
\end{equation*}
\begin{equation*}
{s\mathrm{sutject~to }} \quad  \mathcal{V}^{(k)} = \sum_{i = 1}^{m} \alpha_{i}\mathcal{X}_{i} \quad ( k = 1,\ldots,K,) 
\end{equation*}
\begin{equation}
\quad \sum_{i=1}^{m}\alpha_{i} = 0
\label{eq:6},
\end{equation}
where $\boldsymbol{\alpha} = (\alpha_{1},\ldots,\alpha_{m})^{\top} \in \mathbb{R}^{m}$
are dual variables corresponding to the training data set
$(\mathcal{X}_{i}, y_{i}), i = 1,\ldots,m$,
$D(-\boldsymbol{\alpha})$ is the \emph{conjugate loss function} defined as
\begin{equation*}
D(-\boldsymbol{\alpha}) = \sum_{i=1}^{m} \frac{1}{2}\alpha_{i}^{2} - \alpha_{i}y_{i}
\end{equation*}
in the case of regression with the squared loss \citep{0911.4046v3}, and 
\begin{equation*}
D(-\boldsymbol{\alpha}) = \sum_{i=1}^{m} y_{i}\alpha_{i}\log(y_{i}\alpha_{i}) + (1 - y_{i}\alpha_{i})\log(1 - y_{i}\alpha_{i})
\end{equation*}
with constraint $0 \leq y_{i}\alpha_{i} \leq 1$ in the case of classification with
the logistic loss \citep{0911.4046v3}.
$\delta_{\lambda_{k}}$ is the indicator function defined as $\delta_{\lambda_{k}}(V) = 0$  if $\|V\|_{\mathrm{op}} \leq \lambda_{k}$ and $\delta_{\lambda_{k}}(V) = \mathrm{\infty}$ otherwise.
The constraint $\quad \sum_{i=1}^{m}\alpha_{i} = 0$  is due to the bias term $b$. Here, the auxiliary variables $\mathcal{V}^{(1)},\ldots,\mathcal{V}^{(N)}$ are introduced to remove the coupling between the indicator functions in the objective function (see Appendix~A for details).

The \emph{alternating direction method of multipliers} (ADMM) \citep{Gabay197617,Boyd:2011:DOS:2185815.2185816} has been previously used to solve primal problems of  tensor decomposition  \citep{conf/nips/TomiokaSHK11}   and   multilinear multi-task learning \citep{conf/icml13/multilmtl} with the overlapped  trace norm regularization .  
Optimization in the dual for tensor decomposition problems with the latent and scaled latent trace norm regularization has been solved using ADMM in \citet{conf/nips/TomiokaSHK11}.  
Here, we also adopt ADMM to solve \eqref{eq:6}, 
and describe the formulation and the optimization steps in detail. 

With introduction of dual variables $ \mathcal{W}^{(k)} \in \mathbb{R}^{n_{1}\times \cdots \times n_{K}}, \; k = 1,\ldots,K$ (corresponding to the primal variables of \eqref{eq:5}),  $b \in \mathbb{R}$,
and parameter $\beta > 0$, the augmented Lagrangian function for \eqref{eq:6} is defined as follows:
\begin{multline*}
L(\boldsymbol{\alpha}, \{\mathcal{V}^{(k)}\}_{k=1}^{K}, \{\mathcal{W}^{(k)}\}_{k=1}^{K} , b) 
\\
 =  D(-\boldsymbol{\alpha}) +  \sum_{k =1}^{K} \bigg( \delta_{\lambda_{k}}(V_{(k)}^{(k)}) 
  +  \bigg\langle W_{(k)}^{(k)} ,  \sum_{i = 1}^{m} \alpha_{i}X_{i(k)} - V_{(k)}^{(k)}  \bigg\rangle 
  \\
   +  \frac{\beta}{2} \bigg\|  \sum_{i = 1}^{m} \alpha_{i}X_{i(k)} - V_{(k)}^{(k)}   \bigg\|^{2}_{{\mathrm{F}}}  \bigg) 
 + b\sum_{i=1}^{m}\alpha_{i} + \frac{\beta}{2} \bigg\|\sum_{i=1}^{m}\alpha_{i} \bigg\|^{2}_{\mathrm{F}}.  
\end{multline*}
This ADMM formulation is solved for variables $\boldsymbol{\alpha}$, $\mathcal{V}^{(1)}, \ldots, \mathcal{V}^{(k)}$, $\mathcal{W}^{(1)}, \ldots, \mathcal{W}^{(k)}$, and $b$ 
by considering sub-problems for each variable.
Below, we give the solution for each variable at iterative step $t+1$. 

The first sub-problem to solve is for $\boldsymbol{\alpha}$ at step $t+1$:
\begin{equation*}
\boldsymbol{\alpha}^{t+1} = \underset{\boldsymbol{\alpha}} {\mathrm{argmax}} \,  L(\boldsymbol{\alpha}, \{\mathcal{V}^{(k)t}\}_{k=1}^{K}, \{\mathcal{W}^{(k)t}\}_{k=1}^{K} , b^{t}),
\end{equation*}
where $\{\mathcal{V}^{(k)t}\}_{k=1}^{K}, \{\mathcal{W}^{(k)t}\}_{k=1}^{K}$, and $b^{t}$ are
the solutions obtained at step $t$.  

Depending on the conjugate loss $D(-\boldsymbol{\alpha})$,
the solution for $\boldsymbol{\alpha}$ differs. 
In  the case of regression with the squared loss \eqref{eq:lA},
the  augmented Lagrangian can be minimized with respect to $\boldsymbol{\alpha}$
by solving the following linear equation:
\begin{equation*}
(K\beta\bar{X}\bar{X}^{\top} + I + \beta \mathbf{1}_{m}\mathbf{1}_{m}^{\top})\boldsymbol{\alpha}^{t+1} = (\mathbf{y} - \bar{X}\mathrm{vec}(\mathcal{\bar{W}}^{t}) + \beta \bar{X}\mathrm{vec}(\bar{\mathcal{V}}^{t}) - \mathbf{1}_{m}b^{t}),
\end{equation*} 
where $\bar{X} = [\mathrm{vec}(\mathcal{X}_{1})^{\top}; \cdots;\mathrm{vec}(\mathcal{X}_{m})^{\top} ] \in \mathbb{R}^{m \times N}$,  $\bar{\mathcal{V}}^{t} = \sum_{k=1}^{K}\mathcal{V}^{(k)t}$ , $\bar{\mathcal{W}}^{t} = \sum_{k=1}^{K}\mathcal{W}^{(k)t} $, $\mathbf{y} = (y_{1}, \ldots,y_{m})^{\top} $,
and $\textbf{1}_{m}$ is the $m$-dimensional vector of all ones. Note that, in the above system of equations,
coefficient matrix multiplied with $\boldsymbol{\alpha}$ does not change during optimization.
Thus, it can be efficiently solved at each iteration by precomputing the Cholesky factorization of the matrix.

For classification with the logistic loss \eqref{eq:lB},
the Newton method is used to find the solution for $\boldsymbol{\alpha}^{t+1}$,
which requires the gradient and the Hessian of
$L(\boldsymbol{\alpha},\{\mathcal{V}^{(k)}\}_{k=1}^{K},\{\mathcal{W}^{(k)}\}_{k=1}^{K}, b)$:
\begin{eqnarray*}
\frac{\partial L(\boldsymbol{\alpha}, \{\mathcal{V}^{(k)}\}_{k=1}^{K} , \{\mathcal{W}^{(k)}\}_{k=1}^{K} , b) }{\partial \alpha_{i} } &=& y_{i}\log\bigg(\frac{y_{i}\alpha_{i}}{1 - y_{i}\alpha_{i}}\bigg) + \sum_{k=1}^{K} \langle \mathcal{W}^{(k)t}, \mathcal{X}_{i} \rangle
\\ && +  \beta \sum_{k=1}^{K}  \bigg\langle \mathcal{X}_{i}, \sum_{i = 1}^{m} \mathcal{X}_{i}\alpha_{i}^{t+1} -  \mathcal{V}^{(k)t} \bigg\rangle
 + b + \beta \sum_{i=1}^{m}\alpha_{i}, \\
% \frac{\partial^{2} L(\boldsymbol{\alpha}, \{\mathcal{V}^{(k)}\}_{k=1}^{K} , \{\mathcal{W}^{(k)}\}_{k=1}^{K} , b) }{\partial \alpha_{i}^{2} } &=& \frac{1}{y_{i}\alpha_{i}(1 - y_{i}\alpha_{i})} +   K\beta \langle \mathcal{X}_{i},  \mathcal{X}_{i}   \rangle + \beta  ,\\
\frac{\partial^{2} L(\boldsymbol{\alpha}, \{\mathcal{V}^{(k)}\}_{k=1}^{K} , \{\mathcal{W}^{(k)}\}_{k=1}^{K} , b) }{\partial \alpha_{i}\partial\alpha_{j} } &=&
\begin{cases}
\displaystyle
\frac{1}{y_{i}\alpha_{i}(1 - y_{i}\alpha_{i})} +   K\beta \langle \mathcal{X}_{i},  \mathcal{X}_{i}   \rangle + \beta  & (i=j),\\[5mm]
K\beta  \langle \mathcal{X}_{i},  \mathcal{X}_{j} \rangle + \beta & (i\neq j).   
\end{cases}
\end{eqnarray*}   

Next, we update $\mathcal{V}^{(k)}$ at step $t + 1$
by solving the following sub-problem:
\begin{eqnarray}
\mathcal{V}^{(k)^{t+1} } &=& \underset{\mathcal{V}^{(k)}} {\mathrm{argmax}} \,  L(\boldsymbol{\alpha}^{t+1}, \mathcal{V}^{(k)}, \{\mathcal{V}^{(j)t}\}_{j \neq k}^{K}, \{\mathcal{W}^{(k)t}\}_{k=1}^{K} , b^{t})
\nonumber\\
&=& {\rm{proj}}_{{\lambda_{k}} }\bigg(\frac{W_{(k)}^{(k)t}}{\beta} +  \sum_{i = 1}^{m} \alpha_{i}^{t+1}X_{i(k)}\bigg), \label{eq:o1}
\end{eqnarray}
where ${\rm{proj}}_{\lambda}(W) = U\mathrm{min}(S,\lambda)V^{T} $ and $W = USV^{\top}$.

Finally, we update the dual variables $\mathcal{W}^{(k)}$  and $b$ at step $t+1$ as
\begin{eqnarray}
W_{(k)}^{(k)t+1} &=& W_{(k)}^{(k)t}  + \beta\bigg( \sum_{i = 1}^{m} \alpha_{i}^{t+1}X_{i(k)} - V_{(k)}^{(k)t+1}  \bigg), \label{eq:o2}
\\
b^{t+1} &=& b^{t}  + \beta\sum_{i=1}^{m}\alpha_{i}^{t+1}.
\end{eqnarray}
Note that step \eqref{eq:o1} and step \eqref{eq:o2} can be combined as
\begin{equation*}
W_{(k)}^{(k)t+1} = {\rm{prox}}_{{\beta \lambda_{k}} }\bigg(W_{(k)}^{(k)t} + \beta \sum_{i = 1}^{m} \alpha_{i}^{t+1}X_{i(k)}\bigg), 
\end{equation*}
where ${\rm{prox}}_{\lambda}(W) = U\mathrm{max}(S-\lambda,0)V^{T} $ and $W = USV^{\top}$.
This allows us to avoid computing singular values and the associated singular vectors that are smaller than the threshold $\lambda_{k}$  in \eqref{eq:o1}.

\subsection*{Optimality Condition}
As a stopping condition,
we use the \emph{relative duality gap} \citep{tomioka/techreport1},
which can be expressed as
\begin{equation*}
\frac{P(\mathcal{W}^{t},b^{t}) - D(-\boldsymbol{\hat{\alpha}}^{t})}{P(\mathcal{W}^{t},b^{t})} \leq \epsilon,
\end{equation*}
where $P(\mathcal{W}^{t},b^{t})$ is the primal solution at step $t$ of \eqref{eq:5}
and
$\epsilon$ is a predefined tolerance value.
 $D(-\boldsymbol{\hat{\alpha}}^{t})$ is the dual solution at step $t$ of \eqref{eq:6} with $\boldsymbol{\hat{\alpha}}$ 
%  $\boldsymbol{\alpha}$ (which satisfies $\sum_{i=1}^{m}(\alpha_{i}) \leq \epsilon$) to satisfy the dual feasibility condition $\| V^{(k)}_{(k)} \|_{\mathrm{op}} \leq \lambda_{k}$  and 
obtained by multiplying $\boldsymbol{\alpha}$  with $\min \bigg(1, \frac{\| V(\boldsymbol{\alpha})_{(k)} \|_{\mathrm{op}} }{\lambda_{1}}, \ldots,  \frac{\| V(\boldsymbol{\alpha})_{(k)} \|_{\mathrm{op}} }{\lambda_{K}} \bigg)$, where 
$\mathcal{V}(\boldsymbol{\alpha}) = \sum_{i=1}^{m}X_{i}\alpha_{i}$ 
and
$\|V\|_{\mathrm{op}}$ is the largest singular value of $V$.

\section{Theoretical Risk Analysis}
In this section, we theoretically analyze the excess risk for regularization
with the overlapped trace norm, the latent trace norm, and the scaled latent trace norm.

We consider a loss function $l$ which is Lipshitz continuous with constant $\Lambda$.
Note that this condition is true for both the squared loss and logistic loss functions.
Let the training data set be given as $(\mathcal{X}_{i},y_{i}) \in \mathbb{R}^{n_{1} \times \cdots \times n_{K}} \times \mathrm{Y}, i = 1,\ldots,m$, where $\mathrm{Y} \in \mathbb{R}$ for regression and $\mathrm{Y} \in \{-1,1\}$ for classification. In our theoretical analysis, we assume that elements of $\mathcal{X}_{i}$ independently follow the standard Gaussian distribution.

As the standard formulation \citep{conf/colt13/Andreas},
the empirical risk without the bias term is defined as
\begin{equation*}
\hat{R}(\mathcal{W} ) = \frac{1}{m} \sum_{i=1}^{m} l(\left \langle \mathcal{W}, \mathcal{X}_{i}  \right \rangle , y_{i}),
\end{equation*}
and the expected risk is defined as 
\begin{equation*}
R(\mathcal{W}) = \mathbb{E}_{( \mathcal{X},y) \sim  \mu} l(\left \langle \mathcal{W}, \mathcal{X}  \right \rangle , y),
\end{equation*}
where $\mu$ is the probability distribution from which $(\mathcal{X}_{i},y_{i})$ are sampled.

The optimal $\mathcal{W}^{0}$ that minimizes the expected risk   is given as
\begin{equation}
\mathcal{W}^{0} = \arg\min_{\mathcal{W}} R(\mathcal{W})  \qquad \mathrm{subject \,to\,\,} \|\mathcal{W}\|_{\mathrm{\star}} \leq B_{0}, \label{eq:t1}
\end{equation}
where $\|\cdot\|_{\mathrm{\star}}$ is either
the overlapped trace norm, the latent trace norm, or the scaled latent trace norm. 
The optimal $\mathcal{\hat{W}}$ that minimizes the empirical risk is denoted as
\begin{equation}
\mathcal{\hat{W}} = \arg\min_{\mathcal{W}} \hat{R}(\mathcal{W}) \qquad \mathrm{subject \,to\,\,} \|\mathcal{W}\|_{\mathrm{\star}} \leq B_{0}. \label{eq:t2}
\end{equation}

The next lemma provides an upper bound of the excess risk for tensor-based learning problems (see Appendix~B for its proof), where 
$\|\mathcal{W}\|_{{\mathrm{\star}^{*}}}$ is the dual norm
of $\|\mathcal{W}\|_{{\mathrm{\star}}}$
for ${\mathrm{\star}} = \{\mathrm{overlap}, \mathrm{latent}, \mathrm{scaled} \}$:

\begin{lem} For a given $\Lambda$-Lipchitz continuous loss function $l$ and for any $\mathcal{W} \in \mathbb{R}^{n_{1} \times \cdots \times n_{K}}$ such that $\|\mathcal{W}\|_{{\mathrm{\star}}} \leq B_{0}$ for problems \eqref{eq:t1}--\eqref{eq:t2} , the excess risk for a given training data set $(\mathcal{X}_{i},y_{i}) \in \mathbb{R}^{n_{1} \times \cdots \times n_{K}} \times \mathbb{R}, i = 1,\ldots,m$
is bounded with probability at least $1 - \delta$ as
\begin{equation}
R(\mathcal{\hat{W}}) - R(\mathcal{W}^{0}) \leq \frac{2}{m}\Lambda B_{0} \mathbb{E}\| \mathcal{M} \|_{{\mathrm{\star}}^{*}}  + \sqrt{\frac{\log(\frac{2}{\delta})}{2m}},\label{eq:11}
\end{equation}
where $\mathcal{M} = \sum_{i=1}^{m} \sigma_{i}\mathcal{X}_{i} $ and  $\sigma_{i} \in \{-1 ,1\}$ are Rademacher random variables.
\end{lem}
 
The next theorem gives an excess risk bound for overlapped trace norm regularization
(its proof is also included in Appendix~B),
which is based on
the inequality $\| \mathcal{W} \|_{{\mathrm{overlap}}} \leq \sum_{k=1}^{K}  \sqrt{r_{k}} \| \mathcal{W} \|_{{\mathrm{F}}} $ given in \citet{tomioka/nips13/abs-1303-6370}: 
\begin{thm}
  With probability at least $1-\delta$,
  the excess risk of learning using the overlapped trace norm regularization for any $\mathcal{W}^{0}$ with $\| \mathcal{W}^{0} \|_{\mathrm{F}} \leq B$, multilinear ranks $(r_{1},\dots, r_{K})$, and estimator $\mathcal{\hat{W}}$ with  $B_{0} \leq B\sum_{k=1}^{K}  \sqrt{r_{k}}$  is bounded as
\begin{equation}
R(\hat{W}) - R(W^{0}) \leq c_{1} \Lambda  \frac{B}{\sqrt{m}} \bigg(\sum_{k=1}^{K} \sqrt{r_{k}}\bigg) \min_{k}(\sqrt{n_{k}} + \sqrt{ n_{\backslash k}})  + c_{2} \sqrt{\frac{\log(\frac{2}{\delta})}{2m} },\label{eq:12} 
\end{equation}
where $n_{\backslash k} =  \prod_{j \neq k}^{K}n_{j}$ and $c_{1}$ and $c_{2}$ are constants. 
\end{thm}

In the next theorem, we give an excess risk bound for the  latent trace norm
(its proof is also included in Appendix~B),
which uses the inequality $\|\mathcal{W}\|_{\mathrm{latent}}  \leq \sqrt{\min_{k}r_{k} } \|\mathcal{W}\|_{\mathrm{F}} $ given in \citet{tomioka/nips13/abs-1303-6370}:
\begin{thm}
  With probability at least $1-\delta$,
  the excess risk of learning using the latent norm regularization for any $\mathcal{W}^{0}$ with $\| \mathcal{W}^{0} \|_{\mathrm{F}} \leq B$, multilinear ranks $(r_{1},\dots, r_{K})$, and estimator $\mathcal{\hat{W}}$ with  $B_{0} \leq B\sqrt{\min_{k}r_{k} } $  is bounded as
\begin{multline}
R(\hat{W}) - R(W^{0}) \leq c_{1} \Lambda B \sqrt{\frac{\min_{k}r_{k}}{m}} \bigg(\max_{k}(\sqrt{n_{k}} + \sqrt{ n_{\backslash k}}) + C\sqrt{2\log(K)}  \bigg)
 + c_{2} \sqrt{\frac{\log(\frac{2}{\delta})}{2m} }, \label{eq:13} 
\end{multline}
where  $n_{\backslash k} =  \prod_{j \neq k}^{K}n_{j}$ and ${c}_{1}$, ${c}_{2}$, and $C$ are constants. 
\end{thm}

The above theorem shows that
the excess risk for the latent trace norm \eqref{eq:13} is bounded by the minimum multilinear rank.
If $n_{1} = \cdots = n_{K}$, the latent trace norm is always better then the overlapped trace norm
in terms of the excess risk bounds
because  $\sqrt{\min_{k} r_{k}} < \sum_{k=1}^{K}  \sqrt{r_{k}}$.
If the dimensions $n_{1}, \ldots, n_{K}$ are not the same, 
the overlapped trace norm could be better.
% This fact motives us to closely investigate the scaled latent trace norm.

%Compared with the excess risk for the overlapped trace norm \eqref{eq:12}, when the multilinear ranks are equal or close to each other,  though   $\sqrt{\min_{k} r_{k}} < \sum_{k=1}^{K}  \sqrt{r_{k}}$  the overlapped norm may perform better over the latent trace norm, because  $\sqrt{m}\min_{k}(\sqrt{n_{k}} + \sqrt{ n_{\backslash k}})  \leq \sqrt{m}\max_{k}(\sqrt{n_{k}} + \sqrt{ n_{\backslash k}}) + C\sqrt{2\log(K)} $. However when the multilinear ranks has large variations, the latent trace norm will have a less excess risk since it is bounded by the minimum of multilinear rank. Hence this gives us the theoretical understanding how the latent norm would have the capability to achieve the lowest excess risk with respect to the minimum of ranks. This theoretical finding will be experimentally validated in Section 6.

Finally, we bound the excess risk for the scaled latent trace norm
based on the inequality $\|\mathcal{W}\|_{{\mathrm{scaled}}} \leq \sqrt{\min_{k}\big(\frac{r_{k}}{n_{k}} \big) }\|\mathcal{W}\|_{{\mathrm{F}}}$ given in \citet{nips-14}:
\begin{thm}
  With probability at least $1-\delta$,
  the excess risk of learning using the scaled latent trace norm regularization for any $\mathcal{W}^{0}$ with $\| \mathcal{W}^{0} \|_{\mathrm{F}} \leq B$, multilinear ranks $(r_{1},\dots, r_{K})$, and estimator $\mathcal{\hat{W}}$ with  $B_{0} \leq B\sqrt{\min_{k}\big(\frac{r_{k}}{n_{k}}\big) }$  is bounded as
\begin{multline}
R(\hat{W}) - R(W^{0}) \leq c_{1} \Lambda B \sqrt{\frac{1}{m} \min_{k} \bigg(\frac{r_{k}}{n_{k}}}\bigg) \bigg(\max_{k}(n_{k} + \sqrt{N}) + C\sqrt{2\log(K)}  \bigg) 
 + c_{2} \sqrt{\frac{\log(\frac{2}{\delta})}{2m} }, \label{eq:14}
\end{multline}
where  ${c}_{1}$, ${c}_{2}$, and $C$ are constants. 
\end{thm}

Note that when $n_{1}=\cdots =n_{K}=n$ and the multilinear ranks $r_{1},\ldots, r_{K}$ 
are different, the bounds in Theorem~2 and Theorem~3 are the same. 

Theorem~3 shows that the excess risk for regularization with the scaled latent trace norm
is bounded with the minimum of multilinear ranks relative to their mode dimensions.
Similarly to the latent trace norm, the scaled latent trace norm  would also perform better
than the overlapped norm when the multilinear ranks have large variations.
% because it is the bound with the minimum of  ranks relative to their mode dimensions.
If we consider a ``flat'' tensor,
the modes with small dimensions may have ranks comparable to their dimensions.
Although these modes have the lowest mode-$k$ rank,
they do not impose a low-rank structure. In such cases,
our theory predicts that the scaled latent trace norm performs better
because it is sensitive to the mode-$k$ rank relative to its dimension. 

% If we consider ``flat" tensor structures with modes that are much smaller in dimension  to their other modes, it could commonly occur that the modes with small dimensions may have relative higher ranks compared to their modes with large dimensions. In such cases, the scaled latent trace norm is more accurately bounded with the minimum of relative ranks compared to mode dimensions.     

As a variation, we can also consider a mode-wise ``scaled'' version  of the overlapped trace norm
defined as $\|\mathcal{W}\|_{\mathrm{soverlap}} := \sum_{k=1}^{K} \frac{1}{\sqrt{n_{k} }} \|W_{(k)}\|_{\mathrm{tr}} $. It can be easily seen that
$\| \mathcal{W} \|_{{\mathrm{soverlap}}} \leq \sum_{k=1}^{K}  \sqrt{\frac{r_{k}}{n_{k}}} \| \mathcal{W} \|_{{\mathrm{F}}}$ holds and with the same conditions as in Theorem~1, 
we can upper-bound the excess risk for the scaled overlapped trace norm regularization as
\begin{equation}
R(\hat{W}) - R(W^{0}) \leq c_{1} \Lambda  \frac{B}{\sqrt{m}} \bigg(\sum_{k=1}^{K} \sqrt{\frac{r_{k}}{n_{k}} }\bigg) \min_{k}(n_{k} + \sqrt{N})  + c_{2} \sqrt{\frac{\log(\frac{2}{\delta})}{2m} }.\label{eq:12-2} 
\end{equation}
Note that when all modes have the same dimensions,
$\eqref{eq:12-2}$ coincides with $\eqref{eq:12}$.
Compared with bound  $\eqref{eq:14}$,
the scaled latent norm would perform better than the scaled overlapped norm regularization
since $\min_{k} \sqrt{\frac{r_{k}}{n_{k}}} < \sum_{k=1}^{K} \sqrt{\frac{r_{k}}{n_{k}} }$.

\section{Experiments}
We conducted several experiments using simulated and real-world data
to evaluate the performance of tensor-based regression and classification methods
with regularizations using different tensor norms.
We discuss simulations for tensor-based regression in Section 5.1,
experiments with real-world data for tensor classification in Section 5.2.
For all experiments, we use a MATLAB$^\circledR$ environment
on a 2.10 GHz (2$\times$8 cores) Intel Xeon E5-2450 server machine with 128 GB memory. 

\subsection{Tensor Regression with Artificial Data} 
We report the results of artificial data experiments on tensor-based regression.

We generated three different $3$-mode tensors as weight tensors $\mathcal{W}$ with different multilinear ranks and mode dimensions.
We created two homogenous tensors with equal mode dimensions of $n_{1} = n_{2} = n_{3} = 10$ with different multilinear ranks $(r_{1},r_{2} ,r_{3}) = (3,3,3)$ and $(r_{1},r_{2} ,r_{3}) = (3,5,8)$. 
The third weight tensor is an inhomogenous case with mode dimensions of $n_{1} = 4$, $n_{2} = n_{3} = 10$ and multilinear ranks $(r_{1},r_{2} ,r_{3}) = (3,4,8)$.
To generate these weight tensors, we use the \emph{Tucker decomposition} \citep{journals/siamrev/KoldaB09} of a tensor as
$\mathcal{W}=\mathcal{C} \times_{k=1}^{3} U^{(k)}$,
where $\mathcal{C} \in \mathbb{R}^{r_{1} \times r_{2} \times r_{3}}$ is the core tensor
and $U^{(k)} \in \mathbb{R}^{r_{k} \times n_{k}}$ are component matrices. 
We sample elements of the core tensor $\mathcal{C}$ from a standard Gaussian distribution,
choose component matrices $U^{(k)} \in \mathbb{R}^{r_{k} \times n_{k}}$
to be orthogonal matrices,
and generate $\mathcal{W}$ by mode-wise multiplication of the core tensor and component matrices.
% Using the Tucker decomposition, we first we sample the core tensor of mode dimensions $(r_{1},r_{2} ,r_{3})$ and then multiply this with random orthogonal matrices.

To create training samples $\{\mathcal{X}_{i}, y_{i}\}_{i=1}^{n}$, we first create the random tensors $\mathcal{X}_{i}$ generated with each element independently sampled from the standard Gaussian distribution and obtain $y_{i} = \left \langle \mathcal{W}, \mathcal{X}_{i} \right \rangle + \nu_{i}$,
where $\nu_{i}$ is noise drawn from the Gaussian distribution with mean zero and variance $0.1$. 
In our experiments we use cross validation to select the regularization parameter from range $0.01$--$100$ at intervals of $0.1$. For the purpose of comparison,
we have also simulated matrix regularized regressions for each mode unfolding.
Also, we experimented with cross validation among matrix regularization on each unfolded matrix
to understand whether it can find the correct mode for regularization.
As the baseline vector-based learning method, we use \emph{ridge regression}
(i.e., $l_{2}$-regularized least-squares).

Figure~1 shows the performance of homogenous tensors with equal mode dimensions
$n_{1} = n_{2} = n_{3} = 10$ and equal multilinear ranks $(r_{1},r_{2} ,r_{3}) =$ $(3,3,3)$.
We see that the overlapped norm performs the best,
while both latent norms perform equally (since mode dimensions are equal)
but  inferior to  the overlapped norm.
Also, the regression results from all matrix regularizations with individual modes
perform better than the latent and the scaled latent norm regularized regression models.
Due to the equal multilinear ranks and equal mode dimensions,
it results in equal performance with cross validation among each mode-wise unfolded
matrix regularization.
    
Figure~2 shows the performances of homogenous tensors with equal mode dimensions $n_{1} = n_{2} = n_{3} = 10$ and unequal multilinear ranks $(r_{1},r_{2} ,r_{3}) =(3,5,8)$.
In this case, both the latent and the scaled latent norms also perform equally
since tensor dimensions are the same. 
The mode-$1$ regularized regression models give the best performance 
since it has the lowest rank and regularization
with the latent and scaled latent norms gives the next best performance.
The mode-wise cross validation correctly coincides with the mode-$1$ regularization. 
The overlapped norm performs poorly compared to the latent and  the scaled latent trace norms.

Figure~3 shows the performance of inhomogenous tensors with mode dimensions $n_{1} = 4$, $n_{2} = n_{3} = 10$ and multilinear ranks $(r_{1},r_{2} ,r_{3}) =(3,4,8)$.
In this case, we can see that the scaled latent trace norm outperforms all other tensor norms.
The latent trace norm performs poorly since it  fails to find the mode with the lowest rank.
This well agrees with our theoretical analysis:
as shown in \eqref{eq:13},
the excess risk of the latent trace norm 
is bounded with the minimum of multilinear ranks,
which is on the first mode in the current setup and it is high ranked.
The scaled latent trace norm is able to find the mode with the lowest rank
since it takes the relative rank with respect to the mode dimension as in \eqref{eq:14}.
If we look at the individual mode regularizations,
we see that the best performance is given with the second mode,
which has the lowest rank with respect to the mode dimension,
and the worst performance is given with the first mode,
which is high ranked compared to other modes.
Here, the mode-wise cross validation is again as good as mode-$2$ regularization.

\begin{figure}[p]
\centering
\includegraphics[scale=0.5]{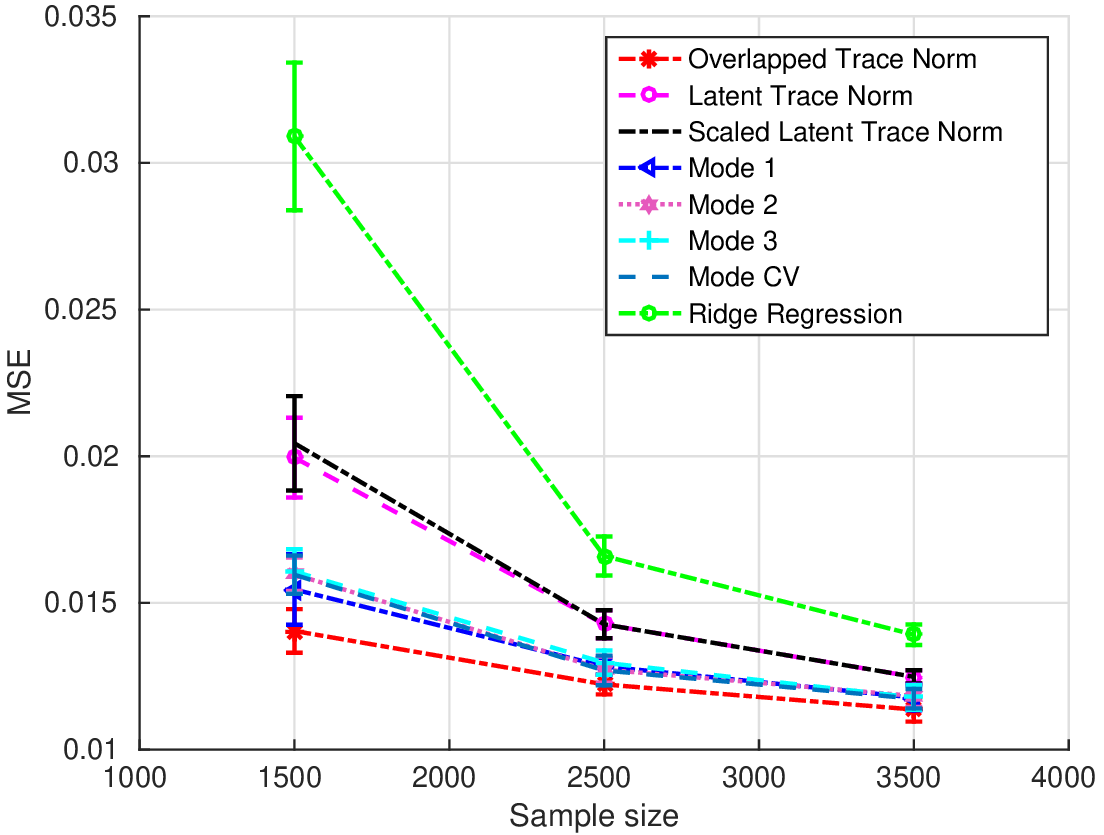}
\label{fig:1}
\caption{Simulation results of tensor regression based on homogenous  weight tensor of equal mode dimensions $n_{1} = n_{2} = n_{3} = 10$ and equal multilinear ranks $(r_{1},r_{2} ,r_{3}) = (3,3,3)$}
% \end{figure}
% \begin{figure}[t]
% \centering
\vspace*{5mm}
\includegraphics[scale=0.5]{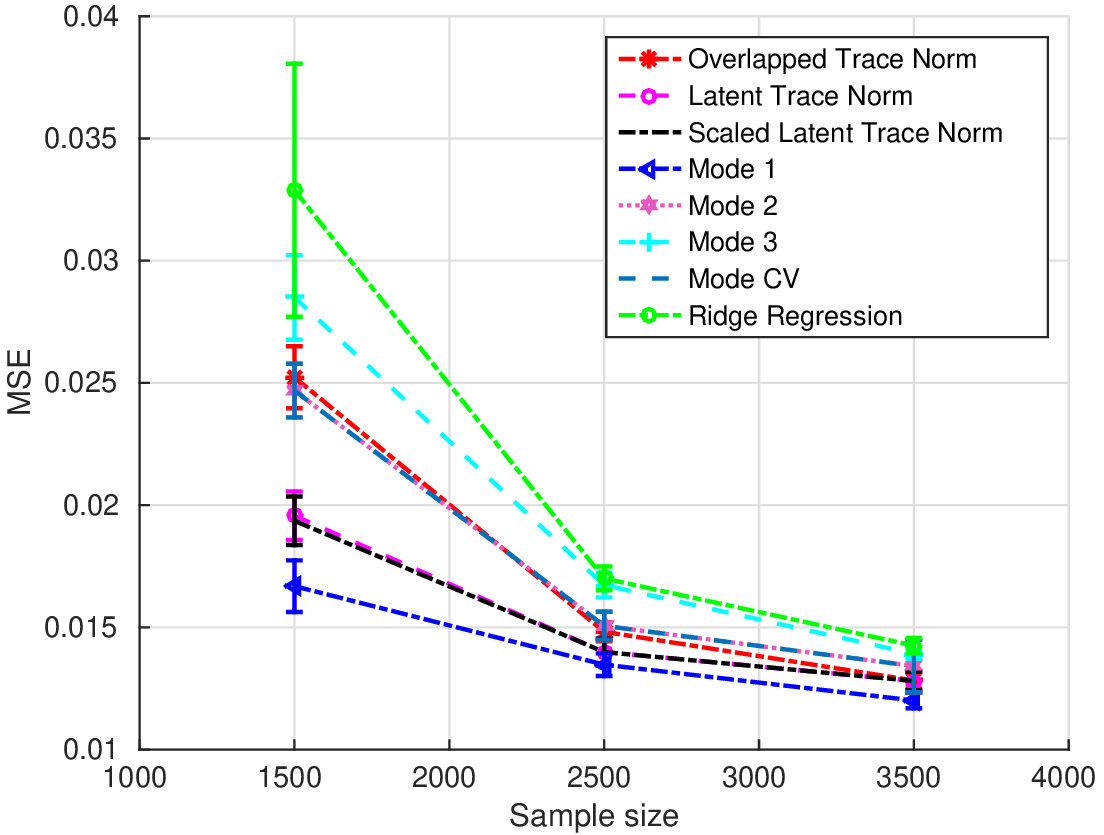}
\label{fig:2}
\caption{Simulation results of tensor regression based on homogenous  weight tensor of equal modes sizes $n_{1} = n_{2} = n_{3} = 10$ and unequal multilinear rank $(r_{1},r_{2} ,r_{3}) = (3,5,8)$}
\end{figure}

\begin{figure}[p]
\centering
\includegraphics[scale=0.5]{toy41010.eps}
\label{fig:3}
\caption{Simulation results of tensor regression based on inhomogenous  weight tensor of equal modes sizes $n_{1} = 4$, $n_{2} = n_{3} = 10$ and  multilinear rank $(r_{1},r_{2} ,r_{3}) = (3,4,8)$}
% \end{figure}
% \begin{figure}[t]
% \centering
\vspace*{5mm}
\includegraphics[scale=0.6]{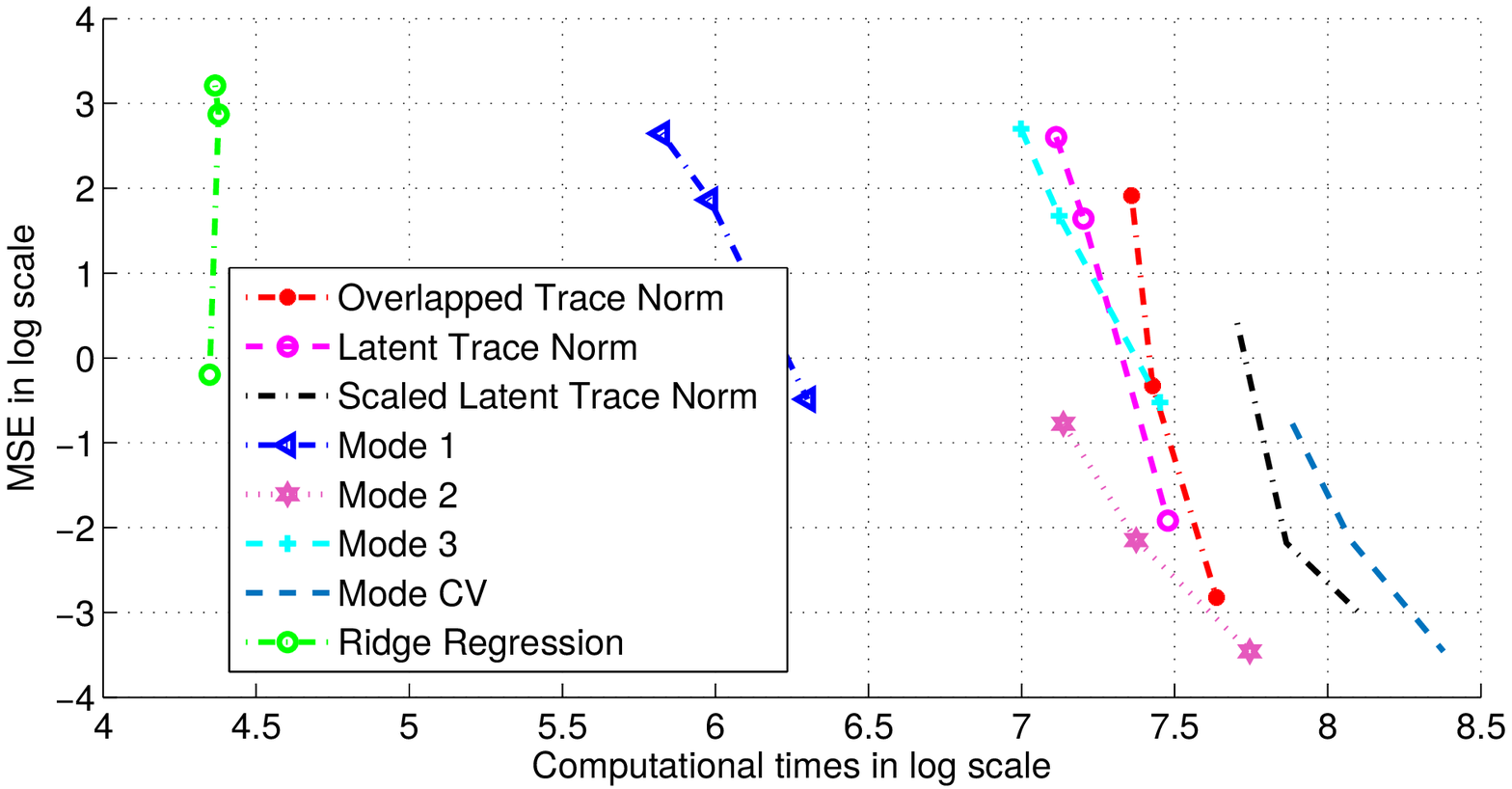}
\label{fig:5}
\caption{Computation times in seconds for toy experiment with inhomogenous tensors with mode dimensions $n_{1} = 4$, $n_{2} = n_{3} = 10$ and multilinear rank $(r_{1},r_{2} ,r_{3}) =$ $(3,4,8)$ }
\end{figure}

It is also worth noticing in all above experiments that ridge regression performed worse than all the tensor regularized learning models. 
This  highlights the necessity of employing low-rank inducing norms
for learning with tensor data without vectorization to get the best performance.  

%In Table \ref{t:1} , we show the computation times for the toy regression experiment with inhomogenous tensors with mode dimensions $n_{1} = 4$, $n_{2} = n_{3} = 10$ and multilinear rank $(r_{1},r_{2} ,r_{3}) =$ $(3,4,8)$ (other experiments show similar behaviour).    

%\begin{table}
%\caption{Computation times in seconds for toy experiment with inhomogenous tensors with mode dimensions $n_{1} = 4$, $n_{2} = n_{3} = 10$ and multilinear rank $(r_{1},r_{2} ,r_{3}) =$ $(3,4,8)$  }    
%\begin{center}   
%    \begin{tabular}{  c | c | c | c |}
%	\cline{2-4}  & \multicolumn{3}{c|}{Training Sample Size } \\ \hline
%
%   \multicolumn{1}{ |c| }{Norm} & 280 & 340 & 400  \\ \hline \hline
%   \multicolumn{1}{ |c| }{Overlapped Trace Norm} & 1571.50(90.23) & 1695.10(235.82) & 2089.50(282.77) \\ \hline
%    \multicolumn{1}{ |c| }{Latent Trace Norm} & 1250.90(171.60)  & 1350.90(171.60)  &  1770.00(201.65) \\ \hline
%    \multicolumn{1}{ |c| }{Scaled Latent Trace Norm} & 2225.70(193.85) & 2610.90(204.51) & 3374.66(391.78)  \\ \hline
%    \multicolumn{1}{ |c| }{Mode 1} & 341.34(46.76) & 398.38(48.82) & 547.47(60.41)  \\ \hline
%    \multicolumn{1}{ |c| }{Mode 2} & 1260.0(92.05) & 1596.80(82.15) & 2317.70(184.04) \\ \hline
%    \multicolumn{1}{ |c| }{Mode 3} & 1095.50(54.12) & 1247.60(155.36) & 1727.70(123.85) \\ \hline
%    \multicolumn{1}{ |c| }{Mode CV}  & 2658.80(123.26) & 3219.30(277.15) & 4359.40(306.10) \\ \hline
%    \multicolumn{1}{ |c| }{Ridge Regression} & 78.76(1.92)  & 79.64(1.40) & 77.31(1.88) \\   \hline
%    \end{tabular}
%\end{center}\label{t:1}
%\end{table} 

Figure~4 shows the computation time for the toy regression experiment
with inhomogenous tensors with mode dimensions $n_{1} = 4$, $n_{2} = n_{3} = 10$ and multilinear ranks $(r_{1},r_{2} ,r_{3}) =$ $(3,4,8)$ (computation time for other setups
showed similar tendency and thus we omit the results). 
For each data set, we measured the computation time of
training regression models,
cross validation for model selection,
and predicting output values for test data.
We can see that methods based on tensor norms  and matrix norms are computationally  much more expensive compared to ridge regression.
However, as we saw above, they achieves higher accuracy than ridge regression.
It is worth noticing that mode-wise cross validation is  computationally more expensive
compared to the scaled latent trace norm and other tensor norms.
This computational advantage  and comparable performance with respect to the best mode-wise
regularization makes the scaled latent trace norm a useful regularization method for tensor-based regression especially for tensors with high variations in its multilinear ranks.

\subsection{Tensor Classification for Hand Gesture Recognition}
Next, we report the results of experiments on tensor classification with the \emph{Cambridge hand gesture data set} \citep{Kim07r.:tensor}. 

The Cambridge hand gesture data set  contains image sequences from  9 gesture classes.
These gesture classes include 3 primitive hand shapes of flats, spread, and V-shape,
and 3 different hand motions of rightward, leftward, and contrast.
Each class has 100 image sequences with  different illumination conditions
and arbitrary motions of two people. Previously, 
the \emph{tensor canonical correlation} \citep{Kim07r.:tensor}
has been used to classify these hand gestures.

To apply tensor classification, first we build action sequences as tensor data
by sampling $S$ images with equal time intervals from each sequence.
This makes  each sequence a tensor of $20 \times 20 \times S$,
where the first two modes are down-sampled images as in \citep{Kim07r.:tensor} 
and $S$ is the number of sampled images.
In our experiments, we set  $S$ at $5$ or $10$. 
We consider binary classification and we have chosen visually similar sequences of
left/flat and left/spread (Figure~5),
which we found to be difficult to classify. 
We apply standardization of data by mean removal and variance normalization.
We randomly sample data into a training set of $120$ data elements,
use a validation set of $40$ data elements to select the optimal regularization parameter,
and finally use a test set of $40$ elements to evaluate the learned classifier.
In addition to the tensor regularized learning models,
we also trained classifiers with matrix regularization with unfolding on each mode separately.
As a baseline vector-based learning method, we have used the $l_{2}$-regularized logistic regression.
We also trained mode-wise cross validation with individual mode regularization (Mode-wise CV).
We repeated the learning procedure for $10$ sample sets for each classifier
and the results are shown in Table~1.    

\begin{figure}[t]
\centering
\includegraphics[scale=0.7]{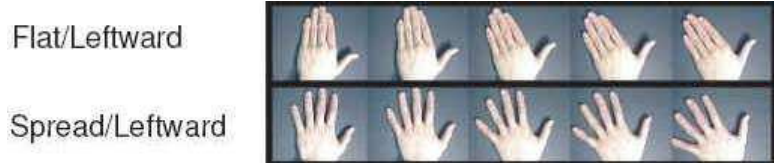}
\label{fig:4}
\caption{Samples of hand motion sequences of left/flat and left/spread}
\end{figure}

\begin{table}[t]
\centering
\caption{Classification error of experiments with the hand gesture data set. The boldfaced figures indicate comparable accuracies among classifiers after a t-test with significance of $0.05$.}    
\label{t:1}
    \begin{tabular}{ c | c | c |}
    \cline{2-3}  & \multicolumn{2}{c|}{Tensor Dimensions } \\ \hline
    \multicolumn{1}{ |c| }{Norm} & (20,20,5) & (20,20,10)  \\ \hline \hline
    \multicolumn{1}{ |c| }{Overlapped Trace Norm} & 0.1425(0.0512) & \textbf{0.0722(0.0363)} \\ \hline
    \multicolumn{1}{ |c| }{Latent Trace Norm} & \textbf{0.1175(0.0487)} & \textbf{0.0806(0.0512)}  \\ \hline
    \multicolumn{1}{ |c| }{Scaled Latent Trace Norm} & \textbf{0.0975(0.0478)} & \textbf{0.0944(0.0512)}  \\ \hline
    \multicolumn{1}{ |c| }{Mode-1} & \textbf{0.1050(0.0422)} & \textbf{0.0950(0.0438)}  \\ \hline
    \multicolumn{1}{ |c| }{Mode-2} & 0.1400(0.0709) & \textbf{0.0900(0.0459)}  \\ \hline
    \multicolumn{1}{ |c| }{Mode-3} & \textbf{0.1200(0.0405)} & 0.1100(0.0592)  \\ \hline
    \multicolumn{1}{ |c| }{Mode-wise CV}  & \textbf{0.1050(0.0542)} & \textbf{0.0950(0.044)}  \\ \hline
    \multicolumn{1}{ |c| }{Logistic regression ($l_{2}$)} & 0.1975(0.0640)  &  0.1925(0.0782) \\   \hline
    \end{tabular}
\end{table}    

In both experiments for $S = 5$ and $10$,
we see that tensor norm regularized classification performs better than
the vectorized learning method. With tensor structure of $(20,20,5)$,
we can see that the scaled latent norm gives the best performance
and the latent trace norm, mode-1, mode-3, and mode-wise cross validation gives
are comparable.
We observed that, with the tensor structure of $(20,20,5)$,
the resulted weight tensor after learning its third mode becomes full rank.
The scaled latent trace norm performed the best since it could identify the mode
with the minimum rank relative to its mode dimension,
which was the first mode in the current setup.
The overlapped trace norm performs poorly due to large variations
in the multilinear ranks and tensor dimensions.

With the tensor structure $(20,20,10)$,
the overlapped trace norm  gives the best performance. 
In this case, we found that the multilinear ranks are close to each other,
which made the overlapped norm to give better performance.
The scaled latent trace norm, latent trace norm, mode-1, mode-2, and mode-wise cross validation
gave comparable performance  with the overlapped trace norm. 
%Again the individual mode regularization performs poorly compared to the best performance given by the overlapped trace norm.        

\subsection{Tensor Classification for Brain Computer Interface}
As our second tensor classification,
we experimented with a motor-imagery EEG classification problem
in the context of \emph{brain computer interface} (BCI).
The objective of the experiments was to classify movements imagined by person 
using the EEG signals captured in that instance.
For our experiments, we used the data from the \emph{BCI competition IVa}
\citep{dornhege2004multiclassbci} . 
Previous research by \citet{conf/icml/TomiokaA07} has considered ``channel $\times$ channel''
as a matrix of the EEG signal and classified it using logistic regression with low-rank matrix regularization. 
Our objective is to model EEG data as tensors to incorporate more information
and learn to classify using tensor regularization methods.   
% \citep{bci4a}
The BCI competition IVa data set consists of BCI experiments of five people. 
Though BCI experiments have used $256$ channels,
we only use signals from $49$ channels following \citet{conf/icml/TomiokaA07}
and pre-process each signal from each channel with $Z$ different band-pass filters
(Butterworth filters).
Let $S_{i} \in \mathbb{R}^{C \times T} $,
where $C$ denotes the number of channels and $T$ denotes the time,
be the matrix obtained by processing with the $i^{\mathrm{th}}$ filter.
As in \citet{conf/icml/TomiokaA07}, each $S_{i}$ is further processed
to make centering and scaling as $\hat{S}_{i}=\frac{1}{\sqrt{T-1}}S_{i}(I_{T}-11^{\top})$.
Then we obtain $X_{i} = \hat{S}_{i}\hat{S}_{i}^{\top}$,
which is a ``channel $\times$ channel'' matrix (in our setting, it is $49 \times 49$).
We arrange all $X_{i} , i =1 ,\ldots,Z$ to form a tensor of dimensions $ Z \times 49 \times 49 $.

For our experiments, we used $Z = 5$ different band-pass Butterworth filters with cutoff frequencies of (7, 10), (9 12), (11 14), (13 16) and (15 18) with scaling by 50 which resulted in a signal converted into a tensor of dimensions $5 \times 49 \times 49$.
We split the data used in the competition into training and validation sets with
proportion of $80:20$, and the rest of the data are used for testing.
As in the previous experiment, we used logistic regression with all the tensor norms,
individual mode unfolded matrix regularizations,
and cross validation with unfolded matrix regularization.
We also used vector-based logistic regression with $l_{2}$-regularization for comparison.
To compare tensor-based methods with the previously proposed  matrix approach \citep{conf/icml/TomiokaA07}, we averaged tensor data over the frequency mode and applied classification with matrix trace norm regularization. For all experiment, we selected all regularization parameters in $100$ splits in logarithmic  scale from  $0.01$ to  $500$.

\begin{sidewaystable}
\small
\centering
\caption{Classification error of experiments with the BCI competition IVa data set.
  The boldfaced figures on the columns \emph{aa}, \emph{al}, \emph{av}, \emph{aw}, and \emph{ay}
  indicate comparable accuracies among classifiers after a t-test with significance of $0.05$. }    
\label{t:2}
    \begin{tabular}{| l | c | c | c | c | c | c |}
        \hline
 	Norm & \shortstack{Subject\\ \emph{aa}} &
        \shortstack{Subject\\ \emph{al}} &
        \shortstack{Subject\\ \emph{av}} &
        \shortstack{Subject\\ \emph{aw}} &
        \shortstack{Subject\\ \emph{ay}} & \shortstack{Avg. Time \\ (Second)}     \\ \hline
    Overlapped Trace Norm 	 	& 0.2205(0.0139) & \textbf{0.0178(0.0)} & \textbf{0.3244(0.0132)} & \textbf{0.0603(0.0071)} & \textbf{0.1254(0.0190)} & 17986(1489) \\ \hline
    Latent Trace Norm 	     	& 0.3107(0.0210) & 0.0339(0.0056)   & 0.3735(0.0218)& 0.1549(0.0381) & 0.4008(0.0) &  20021(14024)\\ \hline
    Scaled Latent Trace  Norm 	&  \textbf{0.2080(0.0043)} & \textbf{0.0179(0.0)}  &  0.3694(0.0182)   & 0.0804(0.0)  & 0.1980(0.0476)  & 77123(149024)  \\ \hline
    Mode-1 					 	& 0.3205(0.0174) & 0.0339(0.0056)  &  0.3739(0.0211) & 0.1450(0.0070)  & 0.4020(0.0038) & 5737(3238)  \\ \hline
    Mode-2 					 	& \textbf{0.2035(0.0124)} & \textbf{0.0285(0.0225)}  & 0.3653(0.0186)  & 0.0790(0.0042)  & 0.1794(0.0025) & 5195(1446) \\ \hline
    Mode-3 					 	& \textbf{0.2035(0.0124)} & \textbf{0.0285(0.0225)}  & 0.3653(0.0186)  & 0.0790(0.0042)  & 0.1794(0.0025) & 5223(1452)  \\ \hline
    Mode-wise CV 				& \textbf{0.2080(0.0369)} & 0.0428(0.0305)  &  0.3545(0.01255)  & 0.1008(0.0227)  & \textbf{0.1452(0.0224)} & 14473(4142) \\ \hline
    Averaged Matrix 			& 0.2732(0.0286) & \textbf{0.0178(0.000)}  &   0.4030(0.2487)  &  0.1366(0.0056)  & 0.1825(0.0)  & 1936(472) \\ \hline
    Logistic regression($l_{2}$)& 0.3161(0.0075)  &  \textbf{0.0179(0.0)}  & 0.3684(0.0537)  & 0.2241(0.0432)  & 0.4040(0.0640) & 72(62) \\   \hline
    \end{tabular}
\end{sidewaystable}

The results of the experiment are given in Table~\ref{t:2},
which strongly indicate that vector-based logistic regression is clearly outperformed
by the overlapped and scaled latent trace norms.
Also, in most cases, the averaged matrix method performs poorly compared to 
the optimal tensor structured regularization methods.
Mode-1 regularization  performs poorly since mode-1 was high ranked compared to the other modes.
Similarly, the latent trace norm gives poor performance since it cannot properly
regularize since it does not consider the rank relative to the mode dimension.
For all subjects, mode-2 and mode-3 unfolded regularizations result in the same performance due to the symmetry of each $X_{i}$ resulting in same rank along mode-2 and mode-3 unfoldings.
For subject \emph{aa}, the scaled latent norm, mode-1, mode-2, and mode-wise cross validation
give the best or comparable performance. In subject \emph{al},
all classifiers except the latent norm and mode-1 regularization gives comparable performance.
For all other subjects except for \emph{aa} and \emph{al},
the overlapped trace norm gives the best performance. 
  
In contrast to the computation time for regression experiments,
in this experiment, we see that the computation time for tensor trace norm regularizations
are more  expensive compared to the mode-wise regularization.
Also, the mode-wise cross validation is computationally less expensive
than the scaled latent trace norm and other tensor trace norms.
This is a slight drawback with the tensor norms,
though they tend to have higher classification accuracy.

\section{Conclusion and Future Work}

In this paper, we have studied tensor-based regression and classification
with regularization using the overlapped trace norm, the latent trace norm,
and the scaled latent trace norm.
We have provided dual optimization methods, theoretical analysis and experimental evaluations
to understand tensor-based inductive learning.
Our theoretical analysis on excess risk bounds showed 
the relationship of excess risks with the multilinear ranks and 
dimensions of the weight tensor.
Our experimental results on both simulated and real data sets
further confirmed the validity of our theoretical analyses.
From the theoretical and empirical results,
we can conclude that the performance of regularization with  tensor norms  depends on the multilinear ranks and mode dimensions,
where the latent and scaled latent norms are more robust in tensors with large variations of multilinear ranks.    
 
Our research opens up many future research directions. For example, an important direction is on improvement of optimization methods. Optimization over the latent tensors that results in the use of the latent  trace norm and the scaled latent trace norm  increases the computational cost compared to  the vectorized methods.   Also, computing multiple singular value decompositions and solving Newton optimization sub-problems (for logistic regression) at each iterative step are computationally expensive. This is evident from our experimental results on computation time for regression and classification.  It would be an important direction to develop computationally more efficient methods for learning with tensor data to make it more practical.

Regularization with a mixture of norms is common in both vector-based (e.g., the \emph{elastic net} \citep{hastieElasticNet}) and matrix-based regularizations \citep{DBLP:conf/icml/SavalleRV12}.
It would be an interesting research direction to combine sparse regularization (the $l_{1}$-norm)
to existing tensor norms. There is also a recent research direction to develop new  composite norms such  the $(k,q)$-trace norm \citep{richardNIPS14}. Development of composite tensor norms can be useful for inductive tensor learning to obtain sparse and low-rank solutions.                 

\section*{Acknowledgment}
KW acknowledges  the Monbukagakusho MEXT Scholarship and KAKENHI 23120004,
RT acknowledges XXXXXX,
and
MS acknowledges the JST CREST program.

\appendix
\section*{Appendix A}
In this appendix, we derive the dual formulation of the latent trace norms.
Let us consider a training data set $(\mathcal{X}_{i}, y_{i}), i = 1,\ldots,m$,
where $\mathcal{X}_{i} \in \mathbb{R}^{n_{1}\times \cdots \times n_{K}}$.  To derive the dual
for the latent trace norms,
we rewrite the primal for the regression of \eqref{eq:1} as
\begin{equation*}
\min_{\mathcal{W}} \sum_{i}^{m} \frac{1}{2}(y_{i} -z_{i})^{2} + \lambda \sum_{k=1}^{K} \| W_{(k)}^{(k)} \|_{{\mathrm{tr}}}
\end{equation*}
\begin{equation*}
{\mathrm{subject~to }} \quad  z_{i} = \left \langle \sum_{k=1}^{K} \mathcal{W}^{(k)},\mathcal{X}_{i} \right \rangle + b, \quad i = 1,\ldots,m.
\end{equation*}
Its Lagrangian  can be written  by introducing variables $\alpha_{i} \in \mathrm{R}, i =1,\ldots,m$ as
\begin{equation*}
\begin{split}
G(\alpha) &= \min_{\mathcal{W},z_{1},\cdots, z_{m}} \sum_{i}^{m} \frac{1}{2}(y_{i} -z_{i})^{2} + \lambda \sum_{k=1}^{K} \| W_{(k)}^{(k)} \|_{{\mathrm{tr}}} + \sum_{i}^{m} \alpha_{i} \bigg( z_{i} - \left \langle \sum_{k=1}^{K} \mathcal{W}^{(k)},\mathcal{X}_{i} \right \rangle + b  \bigg)\\
&= \min_{z_{1},\cdots, z_{m}} \sum_{i}^{m} \bigg( \frac{1}{2}(y_{i} -z_{i})^{2} + \alpha_{i}z_{i} \bigg) + \min_{b}b\sum_{i}^{m}\alpha_{i}\\
&  \hspace{3cm} + \min_{\mathcal{W}} \lambda \sum_{k=1}^{K} \bigg( \| W_{(k)}^{(k)} \|_{{\mathrm{tr}}}  - \left \langle \mathcal{W}_{(k)}^{(k)},\sum_{i}^{m} \alpha_{i}X_{i(k)} \right \rangle \bigg)\\
&=  \sum_{i=1}^{m} \bigg( -\frac{1}{2}\alpha_{i}^{2} + \alpha_{i}y_{i} \bigg) + \sum_{k=1}^{K} \begin{cases}
0 & \| \sum_{i=1}^{m}  \alpha_{i}X_{i(k)} \|_{\mathrm{op}} \leq \lambda_{k} \\
-\infty & \mathrm{otherwise}
\end{cases}\\
& \hspace{8cm} + \begin{cases}
0 &  \sum_{i=1}^{m}  \alpha_{i} = 0 \\
-\infty & \mathrm{otherwise}
\end{cases}\\
&=  \sum_{i=1}^{m} \bigg( -\frac{1}{2}\alpha_{i}^{2} + \alpha_{i}y_{i} \bigg) + \sum_{k=1}^{K}\delta_{\lambda_{k}} \bigg( \sum_{i=1}^{m}  \alpha_{i}X_{i(k)} \bigg) + \delta \bigg( \sum_{i=1}^{m}  \alpha_{i} \bigg).
\end{split}
\end{equation*} 
Let us introduce auxiliary variables $\mathcal{V}^{(1)}, \ldots, \mathcal{V}^{(K)}$ to remove the coupling between the indicator functions.
Then the above dual solutions can be restated as
\begin{equation*}
\min_{\boldsymbol{\alpha}, \mathcal{V}^{(1)},\cdots, \mathcal{V}^{(K)} } \sum_{i=1}^{m} \bigg( -\frac{1}{2}\alpha_{i}^{2} + \alpha_{i}y_{i} \bigg) +  \sum_{k = 1}^{K} \delta_{\lambda_{k}}(V^{(k)}_{(k)})
\end{equation*}
\begin{equation*}
{\mathrm{subject~to }} \quad  \mathcal{V}^{(k)} = \sum_{i = 1}^{m} \alpha_{i}\mathcal{X}_{i} \quad k = 1,\ldots,K, 
\end{equation*}
\begin{equation}
\quad \sum_{i=1}^{m}\alpha_{i} = 0.
\end{equation}
Similarly, we can derive the dual formulation for logistic regression.

\section*{Appendix B}
In this appendix, we prove the theoretical results in Section~4. 

\textit{Proof of Lemma 1}:
{ By using the same approach as the one given in \citet{nips-14,conf/colt13/Andreas},
 we  rewrite 
\begin{multline*}
R(\mathcal{\hat{W}}) - R(\mathcal{W}^{0}) = [R(\mathcal{\hat{W}}) - \hat{R}(\mathcal{\hat{W}})] + [\hat{R}(\mathcal{\hat{W}}) - \hat{R}(\mathcal{W}^{0})] 
 + [\hat{R}(\mathcal{W}^{0}) - R(\mathcal{W}^{0})].
\end{multline*}
The second term is always negative and based on Hoeffding's inequality,
with probability $1- \delta/2$, the third term can be bounded as
$\sqrt{\frac{ln(\frac{1}{\delta})}{2n} }$:
\begin{equation*}
\begin{split}
R(\mathcal{\hat{W}}) - R(\mathcal{W}^{0}) &\leq R(\mathcal{\hat{W}}) - \hat{R}(\mathcal{\hat{W}}) + \sqrt{\frac{\log(\frac{2}{\delta})}{2m} },\\
&\leq \sup_{\|\mathcal{W}\|_{{\mathrm{\star}}} \leq B_{0}}\big(R(\mathcal{W}) - \hat{R}(\mathcal{W}) \big) + \sqrt{\frac{\log(\frac{2}{\delta})}{2m} }.
\end{split}
\end{equation*}
Further applying McDiarmid's inequality, with  probability at least $1-\delta$,
we get the following following Rademacher complexity:
\begin{equation*}
\mathfrak{R} = \frac{2}{m} \mathbb{E} \sup_{\|\mathcal{W}\|_{{\mathrm{\star}}} \leq B_{0}} \sum_{i=1}^{m} \sigma_{i} l(\left \langle \mathcal{W}, \mathcal{X}_{i}  \right \rangle , y_{i}),
\end{equation*}
where $\sigma_{i} \in \{-1,1\}$ are Rademacher variables which leads to 
\begin{equation*}
\begin{split}
R(\mathcal{\hat{W}}) - R(\mathcal{W}^{0}) &\leq \frac{2}{m} \mathbb{E} \sup_{\|\mathcal{W}\|_{{\mathrm{\star}}} \leq B_{0}} \sum_{i=1}^{m} \sigma_{i} l(\left \langle \mathcal{W}, \mathcal{X}_{i}  \right \rangle , y_{i})  + \sqrt{\frac{\log(\frac{2}{\delta})}{2m} }\\
&\leq \frac{2\Lambda}{m} \mathbb{E} \sup_{\|\mathcal{W}\|_{{\mathrm{\star}}} \leq B_{0}} \sum_{i=1}^{m} \sigma_{i} \left \langle \mathcal{W}, \mathcal{X}_{i}  \right \rangle  + \sqrt{\frac{\log(\frac{2}{\delta})}{2m} }\\
&= \frac{2\Lambda}{m}   \mathbb{E}  \sup_{\|\mathcal{W}\|_{{\mathrm{\star}}} \leq B_{0}}\left \langle \mathcal{W},  \sum_{i=1}^{m}  \sigma_{i}  \mathcal{X}_{i}  \right \rangle  + \sqrt{\frac{\log(\frac{2}{\delta})}{2m} }\\
&\leq \frac{2\Lambda}{m}  \mathbb{E} \sup_{\|\mathcal{W}\|_{{\mathrm{\star}}} \leq B_{0}}  \| \mathcal{W}\|_{{\mathrm{\star}}}  \|\mathcal{M} \|_{{\mathrm{\star}}^{*}} + \sqrt{\frac{\log(\frac{2}{\delta})}{2m}} \mbox{~~(H\"older's\;inequality)} \\
&\leq \frac{2\Lambda B_{0}}{m} \mathbb{E}  \|\mathcal{M} \|_{{\mathrm{\star}}^{*}} + \sqrt{\frac{\log(\frac{2}{\delta})}{2m} }. 
\end{split}
\end{equation*}\QEDB
}

\textit{Proof of Theorem 1}:
{First we bound the data-dependent component of $\mathbb{E}\|\mathcal{M}\|_{{\mathrm{overlap}}^{*}}$. For this, we use the following duality relationship borrowed from \citet{tomioka/nips13/abs-1303-6370}:
\begin{equation*}
\| \mathcal{M} \|_{{\mathrm{overlap}}^{*}} = \inf_{\mathcal{M}^{(1)} + \cdots + \mathcal{M}^{(K)} = \mathcal{M}} \max_{k}  \|M_{(k)}^{(k)}\|_{\mathrm{op}}.
\end{equation*}
Since we can take any $\mathcal{M}^{(k)}$ to equal $\mathcal{M}$,
the above norm can be upper bounded as
\begin{equation*}
\|\mathcal{M}\|_{{\mathrm{overlap}}^{*}} \leq \min_{k}  \|M_{(k)}\|_{\mathrm{op}}. \label{eq:20}
\end{equation*}
Furthermore, the expectation of the minimum of $k$ 
can be upper-bounded by the minimum of the expectation:
\begin{equation}
\mathbb{E}\|\mathcal{M}\|_{{\mathrm{overlap}}^{*}} \leq \mathbb{E}\min_{k}  \|M_{(k)}\|_{\mathrm{op}} \leq \min_{k} \mathbb{E} \|M_{(k)}\|_{\mathrm{op}}.\label{eq:21}
\end{equation}
Let $\boldsymbol{\sigma}= \{\sigma_{1},\cdots, \sigma_{m}\}$ be fixed Rademacher variables.
Since each $\mathcal{X}_{i}$ contains elements following the standard Gaussian distribution,
it makes each element in $\mathcal{M}$ a sample from $\mathcal{N} (0,\|\boldsymbol{\sigma}\|_{2}^{2})$.
Based on the standard methods used in \citet{conf/nips/TomiokaSHK11},
we can express $\|M_{(k)}\|_{\mathrm{op}}$ as
\begin{equation*}
\|M_{(k)}\|_{{\mathrm{op}}} = \sup_{u \in \mathit{S}^{n_{k} -1}, v \in \mathit{S}^{\prod_{i \neq k} n_{i} -1}} u^{\top}M_{(k)}v. \label{eq:22}
\end{equation*}
Using Gordan's theorem as in \citet{conf/nips/TomiokaSHK11}, we have
\begin{equation}
\mathbb{E}\|M_{(k)}\|_{{\mathrm{op}}}  \leq  \|\boldsymbol{\sigma}\| \min_{k} (\sqrt{n_{k}} + \sqrt{ n_{\backslash k}}) . \label{eq:23}
\end{equation}
Next taking the expectation over $\sigma$, we have
\begin{equation}
\mathbb{E}\|\boldsymbol{\sigma}\|_{2} \leq \sqrt{\mathbb{E}_{\boldsymbol{\sigma}}\|\boldsymbol{\sigma}\|_{2}^{2}} = \sqrt{m}. \label{eq:24}
\end{equation}
Combining \eqref{eq:23} and \eqref{eq:24} with \eqref{eq:21} results in
\begin{equation*}
\mathbb{E}\|\mathcal{M}\|_{{\mathrm{overlap}}^{*}} \leq \min_{k} \sqrt{m} (\sqrt{n_{k}} + \sqrt{ n_{\backslash k}}).
\end{equation*}
Finally, the excess loss can be written as
\begin{equation*}
R(\hat{W}) - R(W^{0}) \leq c_{1} \Lambda  \frac{B}{\sqrt{m}} \bigg(\sum_{k=1}^{K} \sqrt{r_{k}}\bigg)\min_{k}(\sqrt{n_{k}} + \sqrt{ n_{\backslash k}})  + c_{2} \sqrt{\frac{\log(\frac{2}{\delta})}{2m} }. 
\end{equation*} \QEDB
}

\textit{Proof of Theorem 2} :{To bound the data-dependent component, 
  we use the duality result given in \citet{tomioka/nips13/abs-1303-6370}:
\begin{equation*}
\| \mathcal{M} \|_{{\mathrm{latent}}^{*}} = \max_{k} \| M_{(k)} \|_{\mathrm{op}}.
\end{equation*}
Since $\mathcal{M}$ consists of elements following the standard Gaussian distribution,
for each mode $k$ unfolding, we can write a tail bound \citep{tomioka/nips13/abs-1303-6370} as
\begin{equation*}
P\big(\| M_{(k)} \|_{{\mathrm{op}}} \geq \|\boldsymbol{\sigma}\|(\sqrt{n_{k}} + \sqrt{ n_{\backslash k}}) + t \big) \leq \exp(-t^{2}/(2\sigma^{2})). 
\end{equation*}
Using a union bound, we have 
\begin{equation*}
P\big(\max_{k}\| M_{(k)} \|_{{\mathrm{op}}} \geq \|\boldsymbol{\sigma}\| \max_{k}(\sqrt{n_{k}} + \sqrt{ n_{\backslash k}}) + t \big) \leq K\exp(-t^{2}/(2\sigma^{2})),
\end{equation*}
and this results in 
\begin{equation*}
\mathbb{E}\max_{k}\| M_{(k)} \|_{{\mathrm{op}}} \leq \|\boldsymbol{\sigma}\| \max_{k}(\sqrt{n_{k}} + \sqrt{ n_{\backslash k}}) + \sigma C\sqrt{2\log(K)},
\end{equation*}
where $C$ is a constant.
Similarly to \eqref{eq:24},
taking the expectation over $\boldsymbol{\sigma}$,
we arrive at
\begin{equation*}
\mathbb{E}\max_{k}\| M_{(k)} \|_{{\mathrm{op}}} \leq \sqrt{m} \max_{k}(\sqrt{n_{k}} + \sqrt{ n_{\backslash k}}) + \sqrt{m}C\sqrt{2\log(K)},
\end{equation*}
where $C$ is constant.
Finally, the excess risk is given as
\begin{multline*}
R(\hat{W}) - R(W^{0}) \leq c_{1} \Lambda B \sqrt{\frac{\min_{k}r_{k}}{m}} \bigg(\max_{k}(\sqrt{n_{k}} + \sqrt{ n_{\backslash k}}) + C\sqrt{2\log(K)}  \bigg) 
 + c_{2} \sqrt{\frac{\log(\frac{2}{\delta})}{2m} }. 
\end{multline*}
 \QEDB
}

\textit{Proof of Theorem 3}:
{From \citet{tomioka/nips13/abs-1303-6370}, we have
\begin{equation*}
\| \mathcal{M} \|_{{\mathrm{scaled}}^{*}} = \max_{k} \sqrt{n_{k}}  \| M_{(k)} \|_{\mathrm{op}}. 
\end{equation*} 
Using a similar approach to the latent trace norm with the additional scaling
of $\sqrt{n_{k}}$, we arrive at the following excess bound for the scaled latent trace norm:
\begin{multline*}
R(\hat{W}) - R(W^{0}) \leq c_{1} \Lambda B \sqrt{\frac{1}{m} \min_{k} \bigg(\frac{r_{k}}{n_{k}}}\bigg) \bigg(\max_{k}(n_{k} + \sqrt{N}) + C\sqrt{2\log(K)}  \bigg)
 + c_{2} \sqrt{\frac{\log(\frac{2}{\delta})}{2m} }. 
\end{multline*} 
\qed
}

% \bibliography{norm}

\begin{thebibliography}{}

\bibitem[\protect\astroncite{Bahadori et~al.}{2014}]{NIPS2014_5429}
Bahadori, M.~T., Yu, Q.~R., and Liu, Y. (2014).
\newblock Fast multivariate spatio-temporal analysis via low rank tensor
  learning.
\newblock In 
% Ghahramani, Z., Welling, M., Cortes, C., Lawrence, N., and Weinberger, K., editors, 
{\em Advances in Neural Information Processing Systems 27},
 pages 3491--3499

\bibitem[\protect\astroncite{Bertsekas}{1996}]{1886529043}
Bertsekas, D.~P. (1996).
\newblock {\em Constrained Optimization and Lagrange Multiplier Methods
  (Optimization and Neural Computation Series)}.
\newblock Athena Scientific.

\bibitem[\protect\astroncite{Boyd et~al.}{2011}]{Boyd:2011:DOS:2185815.2185816}
Boyd, S., Parikh, N., Chu, E., Peleato, B., and Eckstein, J. (2011).
\newblock Distributed optimization and statistical learning via the alternating
  direction method of multipliers.
\newblock {\em Foundations and Trends in Machine Learning}, 3(1):1--122.

\bibitem[\protect\astroncite{Cai et~al.}{2010}]{cai2010}
 Cai, J.,  Candès, E. J., and  Shen, Z. (2010).
\newblock A Singular Value Thresholding Algorithm for Matrix Completion.
\newblock {\em SIAM J. on Optimization 20, 4 }, 1956-1982.

\bibitem[\protect\astroncite{Cand\`{e}s
  et~al.}{2011}]{Candes:2011:RPC:1970392.1970395}
Cand\`{e}s, E.~J., Li, X., Ma, Y., and Wright, J. (2011).
\newblock Robust principal component analysis?
\newblock {\em Journal of the ACM}, 58(3):1--37.



\bibitem[\protect\astroncite{Dornhege et~al.}{2004}]{dornhege2004multiclassbci}
Dornhege, G., Blankertz, B., Curio, G., and M\"uller, K.-R. (2004).
\newblock {Boosting bit rates in Noninvasive EEG single-trial classifications
  by feature combination and multiclass paradigms}.
\newblock {\em IEEE Transactions on Biomedical Engineering}, 51(6):993--1002.

\bibitem[\protect\astroncite{Gabay and Mercier}{1976}]{Gabay197617}
Gabay, D. and Mercier, B. (1976).
\newblock A dual algorithm for the solution of nonlinear variational problems
  via finite element approximation.
\newblock {\em Computers \& Mathematics with Applications}, 2(1):17 -- 40.

\bibitem[\protect\astroncite{Gandy et~al.}{2011}]{0266-5611-27-2-025010}
Gandy, S., Recht, B., and Yamada, I. (2011).
\newblock Tensor completion and low-n-rank tensor recovery via convex
  optimization.
\newblock {\em Inverse Problems}, 27(2):025010.

\bibitem[\protect\astroncite{Ganesh et~al.}{2009}]{Lin09fastconvex}
Ganesh, A., Lin, Z., Wright, J., Wu, L., Chen, M., and Ma, Y. (2009).
\newblock Fast convex optimization algorithms for exact recovery of a corrupted
  low-rank matrix.
\newblock In {\em Proceedings of 3rd IEEE International Workshop on
Computational Advances in Multi-Sensor Adaptive Processing}, pages 213--216.




\bibitem[\protect\astroncite{Karatzoglou
  et~al.}{2010}]{Karatzoglou:2010:MRN:1864708.1864727}
Karatzoglou, A., Amatriain, X., Baltrunas, L., and Oliver, N. (2010).
\newblock Multiverse recommendation: N-dimensional tensor factorization for
  context-aware collaborative filtering.
\newblock In {\em Proceedings of the Fourth ACM Conference on Recommender
  Systems}, pages 79--86.

\bibitem[\protect\astroncite{Kolda and Bader}{2009}]{journals/siamrev/KoldaB09}
Kolda, T.~G. and Bader, B.~W. (2009).
\newblock Tensor decompositions and applications.
\newblock {\em SIAM Review}, 51(3):455--500.

\bibitem[\protect\astroncite{Kim et~al.}{2007}]{Kim07r.:tensor}
Kim, T.-K., Wong, S.-F., and Cipolla, R. (2007).
\newblock Tensor canonical correlation analysis for action classification.
\newblock In {\em Proceedings of the {IEEE} Conference on Computer Vision and Pattern Recognition},
pages 1--8.

\bibitem[\protect\astroncite{Liu et~al.}{2010}]{conf/icml/LiuLY10}
Liu, G., Lin, Z., and Yu, Y. (2010).
\newblock Robust subspace segmentation by low-rank representation.
\newblock In
% F{\"u}rnkranz, J. and Joachims, T., editors, 
{\em Proceedings of
  the 27th International Conference on Machine Learning (ICML-10)}, pages
  663--670.

\bibitem[\protect\astroncite{Liu et~al.}{2009}]{DBLP:conf/iccv/LiuMWY09}
Liu, J., Musialski, P., Wonka, P., and Ye, J. (2009).
\newblock Tensor completion for estimating missing values in visual data.
\newblock In {\em Proceedings of the IEEE International Conference on Computer Vision}, pages 2114--2121.

\bibitem[\protect\astroncite{Maurer and Pontil}{2013}]{conf/colt13/Andreas}
Maurer, A. and Pontil, M. (2013).
\newblock Excess risk bounds for multitask learning with trace norm
  regularization.
\newblock In {\em Proceedings of the Annual Conference on Learning Theory 2013}, pages 55--76.

\bibitem[\protect\astroncite{Recht et~al.}{2010}]{doi:10.1137/070697835}
Recht, B., Fazel, M., and Parrilo, P. (2010).
\newblock Guaranteed minimum-rank solutions of linear matrix equations via
  nuclear norm minimization.
\newblock {\em SIAM Review}, 52(3):471--501.

\bibitem[\protect\astroncite{Richard et~al.}{2014}]{richardNIPS14}
Richard, E., Obozinski, G.~R., and Vert, J.-P. (2014).
\newblock Tight convex relaxations for sparse matrix factorization.
\newblock In 
%Ghahramani, Z., Welling, M., Cortes, C., Lawrence, N., and Weinberger, K., editors,
  {\em Advances in Neural Information Processing
  Systems 27}, pages 3284--3292.

\bibitem[\protect\astroncite{Romera{-}Paredes
  et~al.}{2013}]{conf/icml13/multilmtl}
Romera{-}Paredes, B., Aung, H., Bianchi{-}Berthouze, N., and Pontil, M.
  (2013).
\newblock Multilinear multitask learning.
\newblock In {\em Proceedings of the 30th International Conference on Machine
  Learning}, pages 1444--1452.

\bibitem[\protect\astroncite{Sankaranarayanan
  et~al.}{2015}]{10.1371/journal.pone.0121396}
Sankaranarayanan, P., Schomay, T.~E., Aiello, K.~A., and Alter, O. (2015).
\newblock Tensor GSVD of patient- and platform-matched tumor and normal DNA
  copy-number profiles uncovers chromosome arm-wide patterns of tumor-exclusive
  platform-consistent alterations encoding for cell transformation and
  predicting ovarian cancer survival.
\newblock {\em PLoS ONE}, 10(4):e0121396.

\bibitem[\protect\astroncite{Savalle et~al.}{2012}]{DBLP:conf/icml/SavalleRV12}
Savalle, P., Richard, E., and Vayatis, N. (2012).
\newblock Estimation of simultaneously sparse and low rank matrices.
\newblock In {\em Proceedings of the 29th International Conference on Machine
  Learning}, pages 1351--1358.

\bibitem[\protect\astroncite{Signoretto et~al.}{2013}]{SigDinDeLSuy13}
Signoretto, M., Dinh, Q.~T., De~Lathauwer, L., and Suykens, J. A.~K. (2013).
\newblock Learning with tensors: a framework based on convex optimization and
  spectral regularization.
\newblock {\em Machine Learning}, 94(3):303--351.

\bibitem[\protect\astroncite{Tomioka and Aihara}{2007}]{conf/icml/TomiokaA07}
Tomioka, R. and Aihara, K. (2007).
\newblock Classifying matrices with a spectral regularization.
\newblock In 
%Ghahramani, Z., editor,
{\em Proceedings of International Conference on Machine
  Learning}, pages 895--902.

\bibitem[\protect\astroncite{Tomioka et~al.}{2011a}]{tomioka/techreport1}
Tomioka, R., Hayashi, K., and Kashima, H. (2011a).
\newblock {Estimation of low-rank tensors via convex optimization}.
\newblock Technical report, arXiv 1010.0789.

\bibitem[\protect\astroncite{Tomioka and
  Suzuki}{2013}]{tomioka/nips13/abs-1303-6370}
Tomioka, R. and Suzuki, T. (2013).
\newblock {Convex tensor decomposition via structured Schatten norm
  regularization}.
\newblock In {\em Advances in Neural Information Processing Systems 26}, pages 1331--1339.

\bibitem[\protect\astroncite{Tomioka et~al.}{2011b}]{conf/nips/TomiokaSHK11}
Tomioka, R., Suzuki, T., Hayashi, K., and Kashima, H. (2011b).
\newblock {Statistical performance of convex tensor decomposition.}
\newblock
%  In Shawe-Taylor, J., Zemel, R.~S., Bartlett, P.~L., Pereira, F.
%   C.~N., and Weinberger, K.~Q., editors, {\em {NIPS}}, 
\newblock In {\em Advances in Neural Information Processing Systems 24},
pages 972--980.

\bibitem[\protect\astroncite{Tomioka et~al.}{2011c}]{0911.4046v3}
Tomioka, R., Suzuki, T., and Sugiyama, M. (2011c).
\newblock {Super-linear convergence of dual augmented-Lagrangian algorithm for
  sparsity regularized estimation}.
\newblock {\em Journal of Machine Learning Research, 12(May):1537-1586, 2011}.

\bibitem[\protect\astroncite{Wimalawarne et~al.}{2014}]{nips-14}
Wimalawarne, K., Sugiyama, M., and Tomioka, R. (2014).
\newblock Multitask learning meets tensor factorization: task imputation via
  convex optimization.
\newblock In %Ghahramani, Z., Welling, M., Cortes, C., Lawrence, N., and Weinberger, K., editors,
 {\em Advances in Neural Information Processing
  Systems 27}, pages 2825--2833.

\bibitem[\protect\astroncite{Zhou and Li}{2014}]{RSSB:RSSB12031}
Zhou, H. and Li, L. (2014).
\newblock Regularized matrix regression.
\newblock {\em Journal of the Royal Statistical Society: Series B (Statistical
  Methodology)}, 76(2):463--483.

\bibitem[\protect\astroncite{Zhou
  et~al.}{2013}]{doi:10.1080/01621459.2013.776499}
Zhou, H., Li, L., and Zhu, H. (2013).
\newblock Tensor regression with applications in neuroimaging data analysis.
\newblock {\em Journal of the American Statistical Association},
  108(502):540--552.

\bibitem[\protect\astroncite{Zou and Hastie}{2003}]{hastieElasticNet}
Zou, H. and Hastie, T. (2003).
\newblock {Regularization and Variable Selection via the Elastic Net}.
\newblock {\em Journal of the Royal Statistical Society: Series B (Statistical
  Methodology)}, 67(2):301--320.

\end{thebibliography}

\bibliographystyle{apa}

\end{document}